\documentclass[fleqn,12pt]{wlscirep}
\usepackage[utf8]{inputenc}
\usepackage[T1]{fontenc}

\usepackage{graphicx}%
\usepackage{multirow}%
\usepackage{amsmath,amssymb,amsfonts}%
\usepackage{amsthm}%
\usepackage{mathrsfs}%
\usepackage[title]{appendix}%
\usepackage{xcolor}%
\usepackage{textcomp}%
\usepackage{manyfoot}%
\usepackage{booktabs}%
\usepackage{algorithm}%
\usepackage{algorithmicx}%
\usepackage{algpseudocode}%
\usepackage{listings}%
\usepackage{placeins}
\usepackage{xcolor}
\usepackage{float}

\usepackage{soul}

\DeclareMathOperator*{\argmax}{arg\,max}
\DeclareMathOperator*{\argmin}{arg\,min}
\usepackage{enumitem}
\usepackage{bbm}
\usepackage{caption}
\usepackage{graphicx}
\usepackage{hyperref}

\newcommand{\extendcaption}[1]{%
  \refstepcounter{figure}%
  \caption*{\textbf{Extended Data Fig. \arabic{figure}:} #1}%
}
\usepackage{lineno}
\title{
Drift to Remember
}

\author[1]{Jin Du}
\author[2]{Xinhe Zhang}
\author[2]{Hao Shen}
\author[1]{Xun Xian}
\author[1]{Ganghua Wang}
\author[3]{Jiawei Zhang}
\author[1]{Yuhong Yang}
\author[1]{Na Li}
\author[2,*]{Jia Liu}
\author[1,*]{Jie Ding}

\affil[1]{School of Statistics, University of Minnesota Twin Cities, Minneapolis, MN 55455, USA}
\affil[2]{John A. Paulson School of Engineering and Applied Sciences, Harvard University, MA, 02134, USA}
\affil[3]{Dr. Bing Zhang
Department of Statistics, University of Kentucky, Lexington, KY 40506, USA.
}

\affil[*]{jia\_liu@seas.harvard.edu, dingj@umn.edu}

\begin{document}
% \linenumbers

\begin{abstract}
Lifelong learning in artificial intelligence (AI) aims to mimic the biological brain’s ability to continuously learn and retain knowledge, yet it faces challenges such as catastrophic forgetting. Recent neuroscience research suggests that neural activity in biological systems undergoes representational drift, where neural responses evolve over time, even with consistent inputs and tasks. We hypothesize that representational drift can alleviate catastrophic forgetting in AI during new task acquisition. To test this, we introduce DriftNet, a network designed to constantly explore various local minima in the loss landscape while dynamically retrieving relevant tasks. This approach ensures efficient integration of new information and preserves existing knowledge. Experimental studies in image classification and natural language processing demonstrate that DriftNet outperforms existing models in lifelong learning. Importantly, DriftNet is scalable in handling a sequence of tasks such as sentiment analysis and question answering using large language models (LLMs) with billions of parameters on a single Nvidia A100 GPU. DriftNet efficiently updates LLMs using only new data, avoiding the need for full dataset retraining. Tested on GPT-2 and RoBERTa, DriftNet is a robust, cost-effective solution for lifelong learning in LLMs. This study not only advances AI systems to emulate biological learning, but also provides insights into the adaptive mechanisms of biological neural systems, deepening our understanding of lifelong learning in nature.
\end{abstract}

\keywords{bio-inspired AI, lifelong learning, neural representational drift, catastrophic forgetting, adaptive learning systems}

\flushbottom
\maketitle

\thispagestyle{empty}
\clearpage
\section*{Introduction}\label{sec: intro}

\begin{figure}[H]
    \centering    \includegraphics[width=\linewidth]{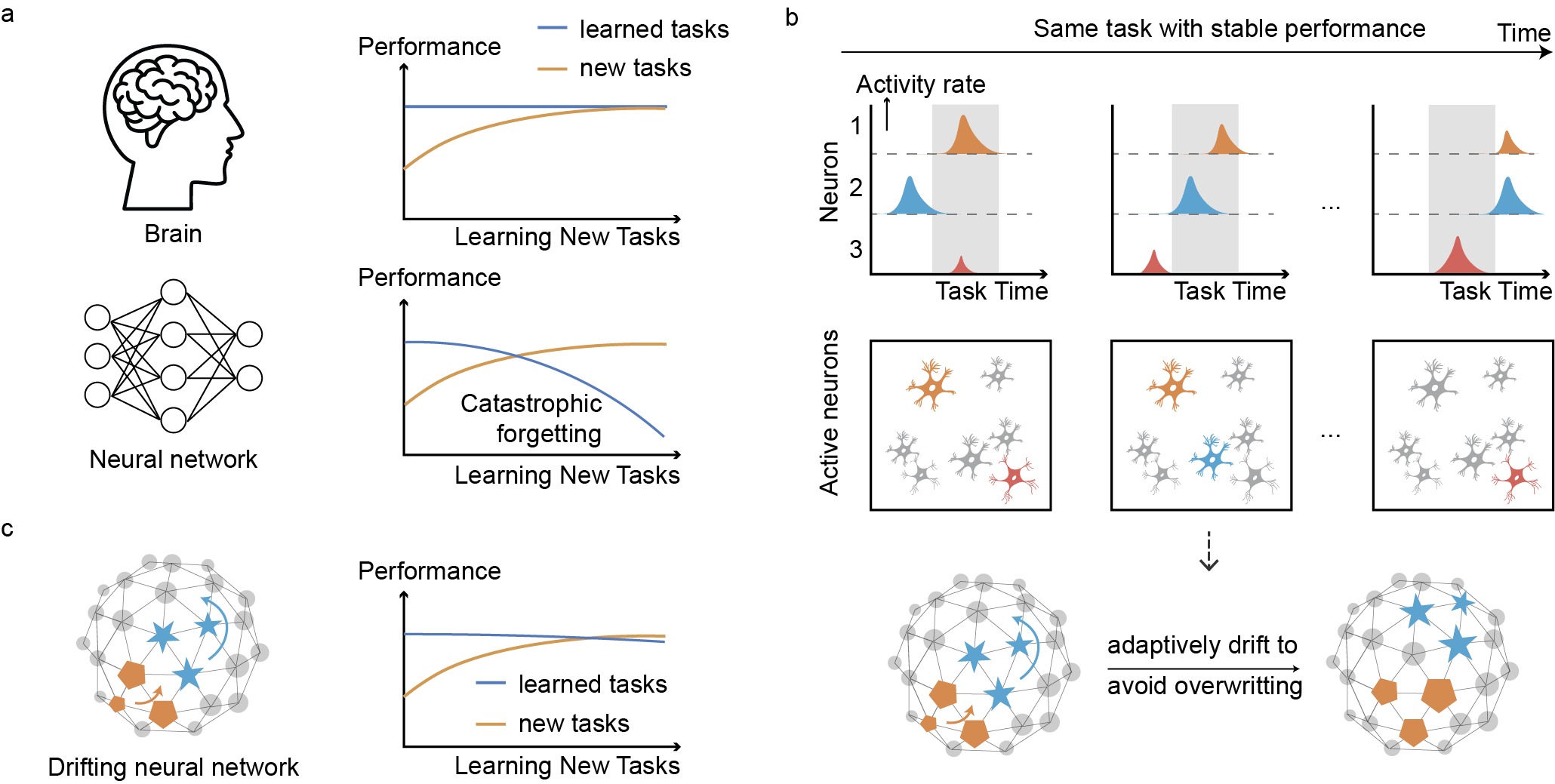}
    \caption{\textbf{Drift and catastrophic forgetting in lifelong learning.}
\textbf{a,} Illustrations of catastrophic forgetting. In a dynamic environment with the continual arrival of new tasks, the brain (top) can learn new tasks while retaining previously acquired knowledge. In contrast, a neural network (bottom) -- when fine-tuned continuously on new tasks -- learns the new tasks but gradually forgets the earlier learned tasks.
\textbf{b,} Schematics of representational drift in biological systems~\cite{driscoll2022representational}. The activity patterns of the same neurons that represent the same task constantly change over time.
\textbf{c,} Illustration of a drift-inspired neural network. The weights of the network drift adaptively over time to avoid overwriting old information while learning new tasks, thus preserving performance on previously learned tasks.
    }
    \label{fig: fig0}
\end{figure}

Biological brains exhibit remarkable lifelong learning skills, acquiring new abilities while retaining previously learned information throughout lifetime. In contrast, this lifelong learning capability, known in artificial intelligence (AI) as continual learning, where a system with limited memory can sequentially learn new tasks without forgetting previous ones, remains a significant challenge. The primary issue is catastrophic forgetting, a phenomenon where the performance in previously learned tasks deteriorates significantly as new tasks are learned~\cite{mccloskey1989catastrophic,parisi2019continual, van2019three} (Figure \ref{fig: fig0}a). This catastrophic forgetting issue limits the lifelong learning capability of current large models, preventing them from evolving over time, especially in applications such as autonomous vehicles, robotics, and natural language processing (NLP). 

To address catastrophic forgetting, current lifelong learning methods fall primarily into three categories: regularization, replay, and architectural methods. Regularization-based methods~\cite{kirkpatrick2017overcoming,zenke2017continual,lee2017overcoming,li2017learning} adjust neural network parameters for new tasks while constraining changes in crucial parameters of previous tasks by imposing constraints on training objectives, such as elastic weight consolidation (EWC)~\cite{kirkpatrick2017overcoming} and synaptic intelligence (SI)~\cite{zenke2017continual}. Replay approaches typically involve training generators for all tasks~\cite{shin2017continual} or maintaining a sample buffer that stores data from previous tasks~\cite{rebuffi2017icarl,rolnick2019experience}. When learning new tasks, data from previous tasks (either as pseudo-samples generated by the generators or as direct samples from the buffer) regularize the training objective. This replay helps to ensure that the performance on previous tasks experiences only minimal degradation. Replayed samples can also prevent gradient updates in crucial directions~\cite{lopez2017gradient,chaudhry2018efficient,guo2020improved}. Architectural strategies allocate new parameters for every task, which can be further divided into two subcategories: (1) fixed architecture, which uses a shared fixed network and trains a distinct set of parameters for every task~\cite{mallya2018packnet,mallya2018piggyback,serra2018overcoming}, and (2) dynamic architecture, which sequentially expands the model structure for new tasks~\cite{rusu2016progressive, aljundi2017expert,yoon2017lifelong}.

However, most existing architectural strategies require task identities during both training and testing phases to be known. Empirical deep learning studies~\cite{mai2022online, van2022three} have demonstrated that while these methods, such as the experience replay (ER)~\cite{rolnick2019experience} and the generative classifier~\cite{van2021class}, perform well on simpler tasks involving classify datasets such as MNIST~\cite{lecun1998gradient} and CIFAR-100~\cite{krizhevsky2009learning} (with a feature extractor pre-trained in CIFAR-10),  they struggle with more challenging tasks such as those involving Mini-ImageNet~\cite{le2015tiny} and CIFAR-100 without pre-trained information. This difficulty is due to the increasing complexity of the data distribution and the higher dimensionality. Empirical experiments~\cite{mai2022online, van2022three} showed that regularization- and replay-based methods that train a single large network face difficulties in encoding new information without compromising existing knowledge.

This raises a fundamental question: What features of biological brains enable them to efficiently encode new information, retain previous knowledge, and effectively recall relevant information upon recurrence of a learned task? Although exact mechanisms remain unclear, recent biological research suggests that even as animals receive the same sensory input and maintain consistent performance on a task, their neural responses can undergo significant drift over time -- a phenomenon termed neural representational drift~\cite{driscoll2017dynamic,kentros2004increased} (Figure \ref{fig: fig0}b). This phenomenon, once considered mainly as measurement artifacts, has been repeatedly confirmed by numerous long-term stable measurements in multiple regions of the brain enabled by advanced measurement techniques~\cite{ziv2013long,marks2021stimulus, rubin2015hippocampal, schoonover2021representational, deitch2021representational, zhao2024realigning}.

Here, we propose that introducing drift in artificial neural networks (ANNs) could be a crucial mechanism to enable lifelong learning by reducing catastrophic forgetting of learned tasks (Figure \ref{fig: fig0}c). Although recent biologically inspired network experiments have suggested multiple mechanisms for implementing representational drift in ANNs~\cite{masset2022drifting, qin2023coordinated, aitken2022geometry, pashakhanloo2023stochastic}, these have not been designed to improve lifelong learning capacity. To address this, we hypothesize that the implementation of a drift mechanism encourages an ANN’s weights and associated hidden representations to continuously change, exploring regions of low loss in the loss landscape, resulting in drifting across multiple local minima. This continuous exploration prevents the network from easily overwriting previously learned weights when acquiring new tasks, thereby enhancing lifelong learning capability. In contrast, a stable deep learning network, where the weights stop changing after converging, ceases to explore different local minima in its loss landscape, leading to a single minimum. When learning a new task, the newly learned weights can overwrite previous ones, causing catastrophic forgetting of earlier tasks (Fig.~\ref{fig: fig1}a).

To test our hypothesis, we developed DriftNet, a drift-inspired lifelong learning framework consisting of two key components: an evolving network that continuously explores various local minima driven by external noise, and a knowledge base that organizes these minima into task-specific groups. DriftNet operates through three steps: exploration, encoding, and retrieval. During exploration, external noise (e.g., batch sampling, dropout, gradient noise, input noise) induces drift in the network's weights, leading to the discovery of diverse local minima in the loss landscape. In the encoding step, these minima are organized within the knowledge base as task-specific groups even when their identities are not explicitly known. New minima related to previously learned tasks are added to existing groups, while unrelated minima form new groups without overwriting prior knowledge. During retrieval, when presented with test inputs, DriftNet assesses the uncertainty of predictions from each group to identify the most relevant task-specific group, enabling the recall of learned knowledge. Local minima from relevant tasks produce outputs with minimal uncertainty, while those from irrelevant tasks yield high uncertainty. Therefore, this drift mechanism is essential as it facilitates continuous exploration and acquisition of diverse local minima, capturing the rich characteristics of each task. These characteristics allow DriftNet to preserve existing knowledge and ensure accurate and robust retrieval, thereby enhancing lifelong learning capability in dynamic environments.

We demonstrate the superior lifelong learning performance of DriftNet by conducting benchmark experiments on simulated datasets and two representative data domains in the field of deep learning: image classification and NLP. Our comparisons include a Stable baseline, where a network is continuously fine-tuned, and a fixed number of recent network copies are retained as knowledge to generate predictions for test inputs. However, in DriftNet, selective copies are retained and adaptively chosen to enhance prediction accuracy.
On simulated data, DriftNet achieves an average test loss of $(1.01 \pm 0.07) \times 10^{-2}$, statistically significantly lower than the Stable baseline $4.22 \pm 0.15$. For image classification tasks, DriftNet achieves an average test accuracy of $80.19 \pm 0.67\%$ on CIFAR-10 and $41.83 \pm 0.75\%$ on CIFAR-100, compared to the Stable baseline $19.18 \pm 0.02\%$ and $12.84 \pm 0.07\%$. In NLP, involving a pre-trained model of $125$ millions of parameters, DriftNet achieves an average test accuracy of $70.37  \pm 1.22\%$, substantially outperforming the Stable baseline $18.29 \pm 0.06\%$. This results demonstrate the robustness, scalability, and wide applicability of drift-inspired models in lifelong learning.

\section*{Results}\label{sec: results}
\subsection*{DriftNet: a drift-inspired lifelong learning framework}\label{sec: drift system}
Figure~\ref{fig: fig1}b presents an overview of DriftNet, which comprises two main components: an evolving model for exploration and a knowledge base for encoding and retrieving grouped task-specific information. DriftNet operates through three main steps: local minima exploration, task encoding, and knowledge retrieval. 

First, to allow the network to actively explore local minima in the loss landscape, we introduced various types of noise into the network (see Methods). These include batch sampling noise from stochastic gradient descent (SGD), dropout~\cite{srivastava2014dropout} (which randomly zeros out nodes), gradient noise~\cite{neelakantan2015adding} (which imposes Gaussian white noise on the gradient), and input noise~\cite{maharana2022review} (which adds Gaussian white noise to the inputs). As the network optimizes the objective function and approaches a local minimum, the introduced noise prompts the network to move away from the current minimum, thus ensuring continuous exploration within the loss landscape (Fig.~\ref{fig: fig1}c). 

Second, during the encoding step, DriftNet employs drift-induced exploration to organize the knowledge base without the need for explicit task identities during training. The knowledge base groups diverse local minima into task-specific groups based on their performance characteristics: local minima from the same task excel at that task but perform differently on unrelated tasks (see Methods). To further reduce memory costs in large language models (LLMs), we employ Parameter-Efficient Fine-Tuning (PEFT)~\cite{houlsby2019parameter, pfeiffer2020adapterhub} strategies, where each local minimum has fewer trainable parameters, such as Low-Rank Adaptation (LoRA)~\cite{hu2021lora} (see Methods).

Third, DriftNet addresses the challenge of efficiently retrieving related knowledge for learned tasks by utilizing encoded knowledge, specifically the task-specific groups of diverse local minima. DriftNet focuses on identifying and retrieving relevant task-specific groups to enhance performance on learned tasks. Each task-specific group in DriftNet, composed of local minima, generates a set of outputs for any given input. The variance of these outputs within each group is quantified as the output uncertainty, which is then used to identify the task identity of any test input. The group with the lowest output uncertainty is selected for retrieval (Fig.~\ref{fig: fig1}e). Various uncertainty measurement methods, including the variance of hard outputs, the variance of soft outputs, and entropy, were explored and detailed in Methods.

In summary, DriftNet continuously explores task-specific loss landscapes, encodes and groups task-specific local minimal in the knowledge space, and uses output uncertainty within each group for retrieval. This drift-induced exploration of loss landscapes provides rich characteristics of task-specific loss landscapes, which is crucial for (1) avoiding overwriting existing knowledge from other tasks during the learning of new tasks and (2) retrieving relevant knowledge when queried by test inputs from learned tasks. Consequently, DriftNet is expected to learn an increasing number of tasks throughout its lifetime, while reducing the forgetting of previous tasks, thus achieving a robust lifelong learning capability.

\subsection*{Benchmarking DriftNet's lifelong learning performance using simulated datasets}

To evaluate and understand the lifelong learning performance of DriftNet, we applied it to simulated datasets. We used linear regression tasks for these simulations because (1) their geometry of local minima can be easily described in a closed form and (2) they provide a clear understanding of how DriftNet functions. In these simulations, the input variables $(x_1,x_2,x_3)$ were used with the output defined as $y=\beta_0^*+\beta_1^* x_1+\beta_2^* x_2+\beta_3^* x_3+\varepsilon$, with $\varepsilon \sim \mathcal{N}(0,0.01)$ representing Gaussian noise. The input covariance matrix was set to be singular, leading to multiple minima (see Methods). %

We trained DriftNet using SGD with Gaussian white noise (mean $0$ and variance $\sigma^2 I_4$) injected into the gradient during training, where $I_4$ is a $4\times4$ identity matrix. Model weights were saved at the end of each epoch{, defined as a single pass through the entire training dataset.}
Every $10$ epochs, the DBSCAN clustering algorithm~\cite{ester1996density} was used to group stored local minima into task-specific groups. During retrieval, for any test input, DriftNet selected the task-specific group with minimal output variance to provide the average output. In contrast, the stable baseline as a control was trained with SGD without noise injection. The weights were stored at each epoch, and the average output of all stored weights was used for any test input. Experiments were conducted with 50 repetitions for the simulated datasets. After learning both tasks, the test loss of DriftNet averaged over two tasks is $(1.01 \pm 0.07) \times 10^{-2}$ when noise level $\sigma=3$, which is significantly lower than the Stable baseline's average test loss of $4.22 \pm 0.15$. These results demonstrate that DriftNet continuously learns both tasks without forgetting.

To understand how drift contributes to avoiding forgetting, we first validated that noise drives the exploration of local minima, which leads to drift. We found that with Gaussian noise injection, the training loss remained steady for both tasks when noise levels ranged from $0$ to $10$ (Fig.~\ref{fig: fig2}d).Meanwhile, the model weights of DriftNet continued to change during learning, actively exploring the task-specific manifold of local minima (Fig.~\ref{fig: fig2}e). Across experiments with various noise levels, the drift rate increased proportionally with the noise level (Extended Data Fig.~\ref{fig: fig2_extended}a). Secondly, we demonstrated that drift-induced task-specific groups of local minima could effectively identify whether any test input was from the relevant task (in-distribution) or irrelevant tasks (out-distribution) based on uncertainty quantification (Fig.~\ref{fig: fig2}h). The results indicated a statistically significant lower uncertainty for in-distribution test inputs, with values of $(7.47\pm 0.02) \times 10^{-2}$ and $(7.29\pm 0.02) \times 10^{-2}$ for the local minima of tasks 1 and 2, respectively. In contrast, the out-of-distribution test inputs showed higher uncertainty, with values of $((1.88 \pm 0.01) \times 10^{-1}$ and $(3.50 \pm 0.02) \times 10^{-1}$ of local minima from Tasks 1 and 2, respectively. The \textit{p}-value was below $0.001$ for all tasks in both the Student's \textit{t}-test (comparing the means of two groups) and the Mann–Whitney \textit{U}-test (comparing the distributions of two groups) (see Methods), suggesting that the uncertainties are statistically different for in-distribution and out-distribution data.
This finding further corroborates that drift enables the accurate retrieval of local minimal in the drifting network. 

We then quantitatively assessed DriftNet's accuracy in retrieving the relevant task-specific group of local minima. The retrieval accuracy increased from $49.98\pm 0.09 \%$ (near the random guess $50\%$) to $94.36 \pm 1.79 \%$ as the noise level increased from $0.001$ to $0.3$, remaining high for noise levels ranging from $0.3$ to $6$, but then decreased to $49.09\pm 0.81 \%$ at a noise level of $10$.  

To understand how DriftNet can continuously group local minima without knowing the task identities, we visualized the performance vectors of local minima using their first two principal components (PCs) (see Methods). The results showed that the performance vectors were well-separated (Fig.~\ref{fig: fig2}f). We also used the adjusted rank index (ARI) score~\cite{rand1971objective} (see Methods) to quantitatively evaluate the grouping quality. The results indicated a high grouping accuracy (above $0.94 \pm 0.01$ when $\sigma=6$) as long as the injected noise was not excessively large ($\sigma\leq 6$) (Extended Data Fig.~\ref{fig: fig2_extended}b), where the local minima had low training losses (Fig.~\ref{fig: fig2}d). %

Overall, the results indicated that higher noise levels led to a more extensive exploration of the local minima manifold (Fig.~\ref{fig: fig2}d-e). These post-exploration task-specific diverse local minima produced similar outputs (low uncertainty) for relevant task inputs and dissimilar outputs (high uncertainty) for irrelevant task inputs. Consequently, the drift mechanism enabled continuous exploration, producing diverse local minima that helped determine the relevance of task-specific local minima to given inputs. This mechanism was particularly effective in retrieving relevant knowledge during lifelong learning when the injected noise was moderately large ($1\leq \sigma\leq 6$, Fig.~\ref{fig: fig2}a-b), thereby preventing forgetting (Fig.~\ref{fig: fig2}f) and ensuring good performance (Fig.~\ref{fig: fig2}a-c).

\subsection*{DriftNet enhances lifelong learning performance in deep learning}%

To demonstrate DriftNet’s ability to enhance lifelong learning performance in deep learning, we applied it to two image classification datasets, CIFAR-10 and CIFAR-100~\cite{krizhevsky2009learning}, which contain $10$ and $100$ classes, respectively. We used a challenging class-incremental learning scenario~\cite{van2022three}, which requires the algorithm to incrementally learn to classify objects from an increasing number of classes. Specifically, CIFAR-10 and CIFAR-100 were divided into $5$ and $10$ subsets with distinct sets of classes, respectively. 

We used the trace of the variance of soft predictions as the uncertainty measurement in classification tasks. For simulated regression datasets, we used the variance of one-dimensional outputs. However, classification tasks with multidimensional soft probabilities present additional challenges. For CIFAR-10, we compared the variance of hard outputs, the variance of soft outputs, and the entropy of average soft outputs (see Methods). The trace of the variance of the soft outputs consistently performed the best in terms of average test accuracy across all noise types and datasets (Extended Data Fig.~\ref{fig: fig3_extended_2}b). %

We applied DriftNet implemented with different sources of noise, including batch sampling noise from SGD, dropout, and additive Gaussian noise applied to gradients and inputs. Baseline methods including Fine-tune (sequentially fine-tuning a network and using it for evaluation), Joint (training on data from all tasks simultaneously in an offline manner to establish the performance upper bound), Theoretical Limits (training and testing with known task identities to achieve the upper bound with optimal retrieval), and Stable (sequentially fine-tuning a network and retaining a fixed number of copies of recent network parameters for evaluation) and state-of-the-art lifelong learning algorithms including Experience Replay (ER)~\cite{rolnick2019experience} and Generative Classifier~\cite{van2021class} were used for comparison. All experiments in this section were conducted for $10$ repetitions. %

DriftNet achieved an average test accuracy of $80.19 \pm 0.67 \%$ (mean $\pm$ SE, $n=10$ throughout the section unless otherwise specified) for CIFAR-10 and $41.83 \pm  0.75 \%$ for CIFAR-100, which was close to the Joint baseline of $84.34 \pm 0.10 \%$ and $49.29 \pm 0.25 \%$ for CIFAR-10 and CIFAR-100, respectively (Fig.~\ref{fig: fig3}a). In contrast, the stable baseline yielded an average test accuracy of $19.18 \pm 0.02\%$ for CIFAR-10 and $12.84\pm 0.07\%$ for CIFAR-100, with no significant improvement as more tasks were learned, indicating its inability to learn new tasks without forgetting previous ones. Among the other lifelong learning baselines, the Generative Classifier~\cite{van2021class} achieved the highest average test accuracy, with $58.67\pm 1.77\%$ for CIFAR-10 and $25.22\pm 0.43\%$  and CIFAR-100 for CIFAR-100.
In summary, DriftNet outperformed all state-of-the-art lifelong learning algorithms and the stable baseline by mitigating forgetting as more tasks are learned in both CIFAR-10 and CIFAR-100 (Extended Data Fig.~\ref{fig: fig3_extended}a). Additionally, DriftNet's performance is close to the Joint baseline (Fig.~\ref{fig: fig3}a).

To investigate whether DriftNet’s ability to alleviate forgetting on CIFAR-10 and CIFAR-100 datasets is due to the drift dynamics of the network, we first examined training loss during the CIFAR-10 task learning with noise levels $\sigma$ ranging from $0$ to $0.01$, with Gaussian white noise (mean $0$, variance $\sigma^2$) injected into the gradient (Fig.~\ref{fig: fig3}b). Representations of the first category, projected onto the first two PCs of drifts (see Methods), changed from the $50$th to the $100$th epoch during the first task when the noise level $\sigma=0.001$ (Fig.~\ref{fig: fig3}c). These results demonstrate that DriftNet maintained steady performance when its parameters kept changing over time, indicating the exploration of various local minima.  %

Next, we tested whether DriftNet can retrieve the relevant task-specific group of local minima. First, we evaluated the uncertainty of each task-specific group across all tasks. On CIFAR-10, the uncertainty of outputs from the local minima of relevant tasks (in-distribution) was relatively low, with values for five tasks ranging from $(8.22 \pm 0.45) \times 10^{-3}$ to $(2.61 \pm 0.07) \times 10^{-2}$. In contrast, the uncertainty of outputs from the local minima of irrelevant tasks (out-of-distribution) was relatively high, with mean values for five tasks concentrating at $0.04$ at a noise level of $\sigma=0.001$ (Fig.~\ref{fig: fig3}f). The results show that the \textit{p}-value is below $0.001$ for all tasks in both the Student's \textit{t}-test and  Mann–Whitney \textit{U}-test (see Methods%
). This indicates that the uncertainty difference is statistically significant for the network in retrieving the specific task from the corresponding group. Second, we assessed the retrieval accuracy. The results show that DriftNet achieved retrieval accuracy with the lowest values of $97.60 \pm 0.34\%$ at $\sigma=0.003$ (Fig.~\ref{fig: fig3}e), with gradient noise ranging from $0$ to $0.03$. These results suggest that this drift-induced exploration enabled the successful retrieval of groups of local minima from relevant tasks.

We further assessed DriftNet’s ability to group local minima without prior knowledge of their task identities. First, the visualization showed that the performance vectors, projected on the first two principal components, were well-separated for CIFAR-10 at a noise level of $0.001$ (Fig.~\ref{fig: fig3}d). Next, we quantitatively evaluated the grouping quality using the ARI~(see Methods). For both datasets and various gradient noise scales, the ARI score was consistently $1$ for all repetitions when the batch size was $16$, indicating that DriftNet can perfectly group stored local minima based on their task identities. These results suggest that stored local minima can be effectively separated by clustering algorithms because outputs from local minima of relevant tasks produced similar results, while those from irrelevant tasks diverged. %

Next, we tested DriftNet's robustness to various noise sources with a batch size of $16$. First, we introduced gradient noise ranging from $\sigma = 0$ to $\sigma = 0.03$. DriftNet's performance remained stable, with values between $80.90\pm 0.47\%$ and $84.96\pm 0.42\%$ (Extended Data Fig.~\ref{fig: fig3_extended}b). Second, we compared different noise sources, including node dropout, gradient noise, input noise, and batch sampling noise (see Methods), using CIFAR-10 and CIFAR-100. The average test accuracy ranged from $80.19\pm 0.67\%$ to $81.04\pm 0.67\%$ for CIFAR-10 and from $41.83\pm 0.75\%$ to $43.95\pm 0.44\%$ for CIFAR-100 (Extended Data Fig.~\ref{fig: fig3_extended_2}a). These results suggest that external noises, including dropout, gradient noise, and input noise, did not significantly affect performance when the batch size was small ($16$). The pairwise Mann-Whitney $U$-test \textit{p}-values between noise types are all above $0.05$, suggesting they are not statistically different. This motivated us to further assess DriftNet with reduced batch sampling noise. %

We then examined the impact of external noise on DriftNet's lifelong learning capability with varying batch sizes. First, we evaluated DriftNet with batch sampling noise only. The results showed that average test accuracy varied slightly from $83.77\pm 0.56\%$ to $83.57\pm 0.39\%$ for small batch sizes ($16$ to $100$), but dropped significantly to $51.87\pm 2.47\%$ when batch sizes increased from $100$ to $3000$ (Extended Data Fig.~\ref{fig: fig3_extended_3}a). This indicates that lifelong learning ability is maintained for small batch sizes, but degrades significantly for larger batch sizes. Second, we investigated whether dropout noise could mitigate the decline in lifelong learning ability as batch sampling noise decreases. With dropout, the average test accuracy decreased from $84.39\pm 0.67\%$ to $60.40\pm 1.95\%$ for batch sizes from $100$ to $3000$ (Extended Data Fig.~\ref{fig: fig3_extended_3}a). The ratio of the average test accuracy with dropout to that with only sampling noise was $1.19\pm 0.07$. These findings suggest that DriftNet maintains its lifelong learning capability under significant noise, whether the batch sampling noise is large (small batch sizes) or small (large batch sizes) with additional dropout noise. Third, we evaluated DriftNet's lifelong learning ability in relation to the severity of noise-induced drift (drift rate) (Extended Data Fig.~\ref{fig: fig3_extended_3}b). The drift rate decreased from $1765.20\pm 76.97$ to $1.69\pm 0.64$ as batch sizes increased with only batch sampling noise. With dropout, the drift rate was higher, decreasing from $2955.41\pm 190.18$ to $24.28\pm 3.03$ as the batch sizes increased from $16$ to $3000$. Furthermore, the results show that the slope of the fitted linear regression line of average test accuracy relative to the logarithm of drift rate is $0.08$, indicating that the average test accuracy increases as the drift rate decreases. {These findings suggest the crucial role of drift induced by externally injected noise, especially when batch sampling noise is reduced, in enhancing lifelong learning capability.}

\subsection*{DriftNet builds effective lifelong learning large language models}
NLP is an important field that aims to enable machines to understand and generate human language~\cite{eisenstein2019introduction, chowdhary2020natural}. Recently, advances in LLMs have shown significant advancements in model architectures, such as Transformer with self-attention~\cite{vaswani2017attention}, and the development of powerful pre-trained language models~\cite{devlin2018bert, achiam2023gpt}. However, training a new language model from scratch for every new task is computationally expensive. For example, training GPT-3, which has $175$ billion parameters, requires $3.14\times 10^{23}$ floating-point operations per second (FLOPS)~\cite{brown2020language}. Under hypothetical conditions without memory limitations, completing this training on a single Nvidia V100 GPU would take around 288 years~\cite{narayanan2021efficient}. On the other hand, naively fine-tuning the pre-trained LLM on a sequence of different language tasks with changing distributions can lead to potential catastrophic forgetting of previously learned knowledge~\cite{sun2019lamol, biesialska2020continual, chen2023lifelong}. 

Given these challenges, we explored whether DriftNet, designed to be compatible with general model architectures and noise sources, could build effective lifelong learning LLMs with limited computational resources. We integrated DriftNet with a pre-trained LLM, GPT-2~\cite{radford2019language}, and sequentially trained it on four language datasets related to topic classification and sentiment analysis: AG's News, Amazon Review Full, DBpedia, and Yahoo! Answers~\cite{zhang2015character} (see Methods).  %
DriftNet achieved an average test accuracy of $70.37  \pm 1.22\%$ (mean $\pm$ SE from $5$ repetitions for this section) over $4$ tasks, while the Theoretical Limits baseline has $80.61 \pm  0.09\%$, the Joint baseline has $80.42 \pm  0.09\%$, and the Stable baseline has $18.29 \pm 0.06\%$ (Fig.~\ref{fig: fig4}b-d). This result demonstrates substantially that DriftNet effectively learns new tasks while mitigating the forgetting of previously learned tasks (Fig.~\ref{fig: fig4}c and Extended Data Fig.~\ref{fig: fig4_extended}b). 

To assess whether drift-induced exploration of various local minima can retrieve related knowledge, we evaluated the uncertainty of outputs from local minima corresponding to relevant and irrelevant tasks, respectively. For all tasks, the uncertainty of outputs from local minima of relevant tasks (in-distribution) is relatively low (ranging from $0.32 \pm 0.04 \times 10^{-4}$ to $1.46 \pm 0.03 \times 10^{-4}$), whereas for irrelevant tasks (out-of-distribution), it is relatively high (ranging from $3.10 \pm 0.04 \times 10^{-4}$ to $7.67 \pm 0.19 \times 10^{-4}$). The uncertainty values are statistically significantly different between the in-distribution and out-of-distribution data, with \textit{p}-values below $10^{-3}$ for all tasks, under both Student's \textit{t}-test and Mann–Whitney \textit{U}-test (see Methods). These results suggest that task-specific groups of local minima exhibit different uncertainty levels in response to related or irrelevant task inputs, allowing the retrieval of relevant knowledge.

\section*{Discussion}
In this paper, we demonstrate that the implementation of drift within artificial neural networks is a critical mechanism to enhance lifelong learning by mitigating catastrophic forgetting. This is achieved through the exploration of diverse local minima in the loss landscape, which helps prevent overwriting and facilitates the future retrieval of previously learned knowledge. Based on this discovery, we introduce a unified general framework, DriftNet, which continuously explores diverse local minima across the loss landscape and encodes them as task-specific groups without overwriting existing tasks. Importantly, DriftNet utilizes uncertainty measures to identify the task-specific group that provides similar outputs with comparatively lower uncertainty. This dynamic process, involving exploration, encoding, and retrieval steps, offers a robust and general lifelong learning solution.
We verified the performance of the DriftNet on simulated datasets, image datasets (CIFAR-10 and CIFAR-100), and NLP tasks using pre-trained LLMs. In these experiments, DriftNet demonstrates superior performance in alleviating the forgetting of learned tasks. Specifically, DriftNet achieved an average test accuracy of $70.37  \pm 1.22\%$ across NLP tasks, significantly higher than the Stable baseline's $18.29 \pm 0.06\%$ and approaching the Theoretical Limits baseline's $80.61 \pm  0.09\%$ and Joint baseline's $80.42 \pm  0.09\%$. Additionally, DriftNet is designed to be highly scalable, reducing computational costs and making it feasible to run on a single Nvidia A100 GPU. While AI has shown impressive capabilities in static environments, its ability to continually adapt to new environments without forgetting acquired skills -- a hallmark of human intelligence -- remains underdeveloped. This adaptability is crucial for autonomous systems like self-driving vehicles and robots, where environments are constantly changing. By incorporating neural representational drift observed in biological brains, DriftNet significantly enhances the lifelong learning capability of artificial neural networks, offering a scalable and adaptable solution applicable across various model structures and AI systems operating in dynamic environments.

We envision several directions for future research. First, further studies on the selection of local minima could improve the effectiveness of uncertainty measures in retrieving relevant knowledge. For example, selecting local minima that are well-separated under certain distances could ensure greater diverse local minima, thereby improving generalization.
Second, DriftNet could be broadly applied in scenarios involving multi-modal data (e.g., vision, language, and audio) or real-time learning systems (e.g., autonomous vehicles and robotics), where continuous learning and decision-making are critical.
Third, hybridizing DriftNet with other lifelong learning approaches could create hybrid models that leverage the strengths of multiple methodologies and make the learning system more robust. For example,
combining DriftNet with reply-based methods
~\cite{rolnick2019experience} to train a shared feature extractor, or with architecture-based methods to train generators, could address challenges related to high dimensionality~\cite{van2021class}.

Our results demonstrate that representational drift can play an important role in the lifelong learning capability of artificial neural networks. However, these results do not necessarily imply that a similar role exists for neural representational drift in biological neural networks. While some hypotheses~\cite{driscoll2022representational, aitken2022geometry} suggest that neural representational drift is critical for the continuous learning capability of the biological brain, substantial biological experiments, particularly precise perturbation experiments, are further required to study the role of neural representational drift in continuous learning in biological brains. Nevertheless, our results
demonstrate that strategically implementing the features identified in the biological brain into artificial neural networks can lead to substantial improvements in machine learning performance.

\clearpage

\section*{Methods}\label{sec: methods}

\subsection*{Definitions}
Let $\mathbb{N}$ denote the set of all positive integers and $\mathbb{R}$ denote the set of all real numbers.
Define $[n] \triangleq \{1, \ldots, n\}$ for any $n\in\mathbb{N}$. For a finite set $A$, let $|A|$ denote its cardinality, namely the number of its elements.
The $\argmax$ of a finite set $A \triangleq \{a_1, \ldots, a_q\}$ is defined as:
\[ \argmax(A) \triangleq \{i \in [q] : a_i = \max(A)\}. \]
The entropy of a vector $\mathbf{a} \triangleq (a_1, \ldots, a_q)$, where $\sum_{i=1}^q a_q=1, a_i >0$ for $i\in [q]$, is defined as:
\[ \text{entropy}(\mathbf{a}) \triangleq -\sum_{i=1}^q a_i \log a_i.\]
The indicator function $\mathbbm{1}(E)$ for any event $E$ is defined as $\mathbbm{1}(E) = 1$ if $E$ occurs and $\mathbbm{1}(E) = 0$ otherwise.
The modulo operation $\bmod$ is defined as $a\bmod b$, which returns the remainder when $a$ is divided by $b$. The dimension of the vector $\omega\in\mathbb{R}^{C}$ ($C\in\mathbb{N}$) is defined as $\text{dim}(\omega)\triangleq C$. For matrix $A$, we denote $A^\top$ as the transpose of $A$.

\subsection*{Lifelong learning formulation}
In this section, we present the mathematical formulation of lifelong learning, which involves a learner sequentially encountering various tasks. Specifically, consider a time interval $[N_i+1, N_{i+1}]$ of learning the $\tau_i$-th task ($\tau_i\in \mathbb{N}, i=1,\ldots$), where $N_i<N_{i+1}\in\mathbb{N}^{+}$ and $\tau_1,\ldots$ are not necessarily different. For any given time $t$ within this interval, the learner processes data consisting of inputs $\mathbf{X}_t \in \mathcal{X}$ and outputs $\mathbf{Y}_t \in \mathcal{Y}$, where the pair $(\mathbf{X}_t, \mathbf{Y}_t)$ is from an underlying data distribution $\mathbb{P}_{\tau_i}$. 
While classical lifelong learning (LL) assumes that transition points ($N_1, N_2, \ldots$) between tasks and task identities ($\tau_1, \ldots$) are known, we consider a more realistic scenario where these are unknown. This situation is more challenging because the learner must infer the current task identity during training, which is particularly difficult when tasks can reoccur or exhibit similar data distributions.  In this paper, we assume the inputs and labels are from $\mathcal{X} = \mathbb{R}^p$ and $\mathcal{Y} = \mathbb{R}$, respectively.

\paragraph{Task}
Assume that the data with inputs $\mathbf{X}$ and labels $\mathbf{Y}$ in a task $\mathcal{T}$ are independent and identically distributed (i.i.d.) with respect to the distribution $(\mathbf{X}, \mathbf{Y}) \sim \mathbb{P}_{\mathcal{T}}$. For any two tasks $\mathcal{T}_1$ and $\mathcal{T}_2$, we consider them identical if and only if $\mathbb{P}_{\mathcal{T}_1}(\mathbf{X}, \mathbf{Y}) = \mathbb{P}_{\mathcal{T}_2}(\mathbf{X}, \mathbf{Y})$ for all $\mathbf{X}\in\mathcal X$ and $\mathbf{Y}\in\mathcal Y$. 
In this paper, we focus on the scenario where each task appears only once, such that $\tau_i = i$ for $i \in \mathbb{N}$.

\paragraph{Test accuracy}
The lifelong learning capability of a learner is evaluated by its ability to learn new tasks without forgetting previously learned ones after training on $k$ tasks.
This is quantified by the average test accuracy in all $k$ tasks, where the test accuracy of the $i$-th task is defined as:
\[
\text{Acc}_i\triangleq
\frac{1}{|\mathcal{D}_{\text{test},i}|}\sum_{(\mathbf{X}_{\text{test}},\mathbf{Y}_{\text{test}})\in \mathcal{D}_{\text{test},i}} \mathbbm{1}(\hat{\mathbf{Y}}_{\text{test}}=\mathbf{Y}_{\text{test}})
,
\]
where $\mathcal{D}_{\text{test},i}$ consists of data from $i$-th task, and $\hat{\mathbf{Y}}_{\text{test}}$ is the model prediction. 

\paragraph{Retrieval accuracy}
During the testing phase, given test inputs $\mathbf{X}_{\text{test}}$ from task $\tau_{\text{test}}$, DriftNet retrieves a set of local minima $\{\mathbf{m}_1, \mathbf{m}_2, \ldots, \mathbf{m}_k\}$  that were trained on tasks $\{\tau_1, \tau_2, \ldots, \tau_k\}$, where $k\in\mathbb{N}^+$. We define the retrieval of relevant knowledge as successful if the majority of the retrieved local minima are relevant to the current task $\tau_{\text{test}}$, expressed as:
\[
r(\mathbf{X}_{\text{test}}, \tau_{\text{test}}) \triangleq \mathbbm{1}\left(\sum_{j=1}^{k} \mathbbm{1}(\tau_j = \tau_{\text{test}}) > \frac{k}{2}\right).
\]
The retrieval accuracy of task $\tau_{\text{test}}$ is the average of $r(\mathbf{X}_{\text{test}}, \tau_{\text{test}})$ over all test inputs $\mathbf{X}_{\text{test}}$ from task $\tau_{\text{test}}$. 
The (overall) retrieval accuracy is the average retrieval accuracy across all tasks.

\subsection*{Adjusted Rand Index (ARI)}\label{sec: ari}

The Adjusted Rand Index (ARI)~\cite{hubert1985comparing} is a measure of the similarity between two data groupings (partitions).
Let $a_i$ $(i=1,\dots, r)$ denote the number of observations from the $i$-th group of the first grouping and  $b_j$ $(j=1,\dots, s)$ denote the number of observations from the $j$th group of the second grouping, where $r,s\in\mathbb{N}$. Let $n_{ij}$  $(i=1,\dots, r, j=1,\dots, s)$ denote the number of observations from both the $i$th group of the first grouping and $j$th group of the second grouping.
ARI is calculated by:
\[
\frac{\sum_{ij} \binom{n_{ij}}{2} - \left[ \sum_i \binom{a_i}{2} \sum_j \binom{b_j}{2} \right] / \binom{n}{2}}{ \frac{1}{2} \left[ \sum_i \binom{a_i}{2} + \sum_j \binom{b_j}{2} \right] - \left[ \sum_i \binom{a_i}{2} \sum_j \binom{b_j}{2} \right] / \binom{n}{2}}.
\]
 ARI takes values between -1 and 1. An ARI of $1$ indicates perfect agreement between the two groupings, $0$ indicates random labeling, and negative values indicate worse than random labeling.

\subsection*{Criterion}

The Mean Squared Error (MSE) loss evaluates regression model performance and is defined as:
\[
\frac{1}{n} \sum_{j=1}^{n} (\mathbf{Y}_j - \hat{\mathbf{Y}}_j)^2,
\]
where $\mathbf{Y}_j$ is the observed (actual) output value for the $j$-th sample, $\hat{\mathbf{Y}}_j$ is its predicted value, and $n$ is the sample size.

Cross-Entropy loss evaluates classification model performance. For multi-class classification with $C$ classes, it is defined as:
\[
- \frac{1}{n} \sum_{j=1}^{n} \sum_{c=1}^{C} \mathbf{Y}_{jc} \log(\hat{\mathbf{Y}}_{jc}),
\]
where $\mathbf{Y}_{jc}$ is a binary indicator (0 or 1) if the class label $c$ is correct for sample $j$, and $\hat{\mathbf{Y}}_{jc}$ is the predicted probability that the sample $j$ belongs to class~$c$.

\subsection*{Student's \textit{t}-test and Mann-Whitney \textit{U}-test}\label{sec: tests}
To determine whether the measurements of the two groups are statistically different, we have employed both Student's \textit{t}-test~\cite{student1908probable} and Mann-Whitney \textit{U}-test~\cite{mann1947test}. Student's \textit{t}-test was used to compare the means of two groups, assuming normally distributed data with equal variances. The t-statistic is calculated as:
\[
t = \frac{\bar{X}_1 - \bar{X}_2}{\sqrt{S_p^2(\frac{1}{n_1}+\frac{1}{n_2})}},
\]
where $S_p^2\triangleq \frac{(n_1-1)S_1^2+(n_2-1)S_2^2}{n_1+n_2-2}$, $\bar{X}_1$ and $\bar{X}_2$ are the sample means, $S_1^2$ and $S_2^2$ are the sample variances, and $n_1$ and $n_2$ are the sample sizes of the two groups, respectively. Under the null hypothesis, the t-statistic follows a t-distribution with $n_1 + n_2 - 2$ degrees of freedom.

The Mann-Whitney \textit{U}-test was used to determine whether two groups have different distributions. 
It ranks all data points from both groups and calculates the $U$ statistic:
\[
U = n_1 n_2 + \frac{n_1(n_1 + 1)}{2} - R_1,
\]
where $R_1$ is the sum of the ranks for the first group, and $n_1$ and $n_2$ are the sample sizes of the first and second groups, respectively. Under the null hypothesis, the $U$ statistic follows a distribution that can be approximated by a normal distribution when the sample size is large.

\subsection*{Low-rank adaptation (LoRA)}\label{sec: lora}
Low-Rank Adaptation (LoRA)~\cite{hu2021lora} is a parameter-efficient technique for fine-tuning large pre-trained models. 
Specifically, for a pre-trained large weight matrix $W_0 \in \mathbb{R}^{m \times n}$, let
$\Delta W$ be its update during the fine-tuning, that is, the updated weighted matrix is
$W_0+\Delta W$.
LoRA constrains each update to have a low-rank representation:
\[
\Delta W = \alpha BA,
\]
where $B \in \mathbb{R}^{m \times r}$ and $A \in \mathbb{R}^{r \times n}$ are low-rank matrices with rank $r \ll \min(m, n)$, and $\alpha > 0$ is a scaling factor. During the entire training stage, the pre-trained weights $W_0$ are fixed while $A$ and $B$ are trainable parameters, thus requiring fewer trainable parameters. 
To produce predictions during the inference stage, the contribution of the low-rank matrices can be integrated into the updated weight matrix: 
\[
W' = W_0 + \alpha BA.
\]

\subsection*{Components of DriftNet}\label{sec: algorithm}
In this section, we elaborate on DriftNet in Algorithm~\ref{alg: driftnet}, a drift-inspired lifelong learning framework. At any given time $t$, DriftNet consists of two main components: (1) an evolving model $M_{\theta_t}$, which is updated with noise to encourage exploration, and (2) a knowledge base $\mathcal{K}$, which stores various local minima $\theta_{n_1}, \ldots, \theta_{n_{m(t)}}$ learned from the evolving model, where the time indices $1\leq n_1<\ldots,n_{m(t)}\leq t$, and $m(t):[0,\infty)\rightarrow \mathbb N$ denote the number of local minima at time $t$. These local minima of different tasks are grouped into task-specific groups, with grouping identities $\mathbf{gr}_t \triangleq (\text{gr}_{t,1}, \ldots, \text{gr}_{t,n_{m(t)}}) \in \mathbb{N}^{m(t)}$, where the $i$-th stored minimum $\theta_{n_i}$ belongs to the $\text{gr}_{t,i}$-th group, for $i\in[m(t)]$. 

DriftNet operates in three key steps, further detailed in the following sections. Here is an overview of Algorithm~\ref{alg: driftnet}:
\begin{enumerate}
    \item \textbf{Exploration Step}~(Lines~\ref{line: alg1 explore start}-\ref{line: alg1 explore end}): The evolving model $M_{\theta_t}$ is updated by injecting designed noise to facilitate the exploration of new solutions.
    \item \textbf{Encoding Step}~(Lines~\ref{line: alg1 enc start}-\ref{line: alg1 enc end}): The current state of the evolving model (treated as a local minimum) is stored for every given time interval. The stored local minima are then clustered into task-specific groups based on their performance, evaluated by a small buffer that stores previous data with equal probability.
    \item \textbf{Retrieval Step}~(Lines~\ref{line: alg1 retrieve start}-\ref{line: alg1 retrieve end}): When given a test input, DriftNet retrieves the output of the task-specific group that exhibits the lowest uncertainty in its output.
\end{enumerate}

\begin{algorithm}
    \caption{DriftNet}
    \begin{algorithmic}[1]
        \renewcommand{\algorithmicrequire}{\textbf{Input:}}
        \renewcommand{\algorithmicensure}{\textbf{Output:}}
        \Require Encode interval $n_{\text{enc}} \in \mathbb{N}$, buffer size $n_B$.
        
        \State Initialize the evolving model $M_{\theta_1}: \mathbb{R}^{n \times p} \mapsto \mathbb{R}^n$, knowledge base $\mathcal{K} = \emptyset$, grouping identities $\mathbf{gr} = \mathbb{N}^{|\mathcal{K}|} = \emptyset$, and buffer $\mathcal{B} = \emptyset$.
        
        \For{$t = 1, 2, \ldots$}
            \State Receive inputs $\mathbf{X}_t$, and labels $\mathbf{Y}_t$.
            \State \label{line: alg1 explore start} 
            \hspace*{-\parindent}~~~~~~~\textbf{Exploration step:}
            \State \label{line: alg1 explore end} $\theta_{t+1} \gets \text{NoisyUpdate}(\theta_t, \mathbf{X}_t, \mathbf{Y}_t, \sigma)$.

            \State \label{line: alg1 enc start} \textbf{Encoding step:}
            \State \label{line: alg1 buffer} $\mathcal{B} \gets \text{BufferUpdate}(\mathcal{B}, \mathbf{X}_t, \mathbf{Y}_t, t)$.
            \If{$t \bmod n_{\text{enc}} = 0$}
                \State $\mathcal{K}, \mathbf{gr} \gets \text{Encode}(\mathcal{K}, \theta_t, \mathcal{B}).$
            \EndIf \label{line: alg1 enc end}

            \State \label{line: alg1 retrieve start} \textbf{Retrieval step:}
            \If{Receive test inputs $\mathbf{x}_{\text{test}}$}
                \State $\hat{y}_{\text{test}} \gets \text{Retrieve}(\mathcal{K}, \mathbf{gr}, \mathbf{x}_{\text{test}}).$
            \EndIf \label{line: alg1 retrieve end}
        \EndFor
    \end{algorithmic}
\label{alg: driftnet}
\end{algorithm}

\begin{algorithm}
\caption{Exploration step}
\begin{algorithmic}[1]
    \renewcommand{\algorithmicrequire}{\textbf{Input:}}
    \renewcommand{\algorithmicensure}{\textbf{Output:}}
    \Require Noise Type $\tt{type}$, learning rate $\eta > 0$,  data $\mathbf{X}\in\mathbb{R}^{n\times p}, \mathbf{Y}^{n}$, and noise scale $\sigma>0$.
        \If{$\tt{type}$ is ``inputs''}
            \State $\mathbf{X} \gets \mathbf{X} + \varepsilon$, where $\varepsilon \sim \mathcal{N}(0, \sigma^2 I_p)$.
        \EndIf
        \State $g \gets \nabla_{\theta} \text{CrossEntropy}(M_{\theta}(\mathbf{X}), \mathbf{Y})$.
        \If{$\tt{type}$ is ``gradient''}
            \State $g \gets g + \tilde{\varepsilon}$, where $\tilde{\varepsilon} \sim \mathcal{N}(0, \sigma^2 I_{\text{dim}(\theta)})$.
        \EndIf
        \State $\theta \gets \theta - \eta g$.
    \Ensure  $\theta$
\end{algorithmic}
\label{alg: explore}
\end{algorithm}

\begin{algorithm}
\caption{Encoding step}
\begin{algorithmic}[1]
    \renewcommand{\algorithmicrequire}{\textbf{Input:}}
    \renewcommand{\algorithmicensure}{\textbf{Output:}}
    \Require Knowledge $\mathcal{K}$, current parameters $\theta_t$, batch data $\mathcal{B} = \{(\tilde{\mathbf{X}}_1, \tilde{\mathbf{Y}}_1), \ldots, (\tilde{\mathbf{X}}_{|\mathcal{B}|}, \tilde{\mathbf{Y}}_{|\mathcal{B}|})\}$.

        \State $\mathcal{K} \gets \mathcal{K} \cup \{\theta_t\}$ 
        \For{every $\tilde{\theta} \in \mathcal{K}$}
        \For{$j=1\to |\mathcal{B}|$}
            \State 
             $pv_{\tilde{\theta},j} \gets \text{CrossEntropy}(M_{\tilde{\theta}}(\tilde{\mathbf{X}}_j), \tilde{\mathbf{Y}}_j).$
        \EndFor
        \EndFor
        \State $\mathbf{gr} \gets \text{Cluster}(\{(pv_{\tilde{\theta},1}, \ldots, pv_{\tilde{\theta},|\mathcal{B}|}) : \tilde{\theta} \in \mathcal{K}\}).$ 
 \Ensure  $\mathcal{K}$, $\mathbf{gr}$
\end{algorithmic}
\label{alg: encoding}
\end{algorithm}

\begin{algorithm}
\caption{Retrieval step}
\begin{algorithmic}[1]
    \renewcommand{\algorithmicrequire}{\textbf{Input:}}
    \renewcommand{\algorithmicensure}{\textbf{Output:}}
    \Require Knowledge base $\mathcal{K}=\{\tilde{\theta}_1, \ldots, \tilde{\theta}_{|\mathcal{K}|}\}$, group labels $\mathbf{gr}=(\text{gr}_1,\ldots,\text{gr}_{|\mathcal{K}|})$, test input $\mathbf{X}_{\text{test}}$
    
    \For{$i = 1$ to $\max{(\mathbf{gr})}$}
        \State $\mathcal{K}_i \gets \{\tilde{\theta}_j : \text{gr}_j = i, j \in [|\mathcal{K}|]\}.$
        
        \State $\mathcal{Y}_i \gets \{M_{\tilde{\theta}}(\mathbf{X}_{\text{test}}) : \tilde{\theta} \in \mathcal{K}_i\}$ 
        \State $u_i \gets \text{UncertaintyMeasure}(\mathcal{Y}_i)$ 
    \EndFor
    \State $i_{\text{min}} \gets \argmin\{u_1, \ldots, u_{\max{(\mathbf{gr})}}\}$
    \State $\hat{y}_{\text{test}} \gets \text{Mean}(\mathcal{Y}_{i_{\text{min}}})$
\end{algorithmic}
\label{alg: retrieve}
\end{algorithm}

\begin{algorithm}
\caption{BufferUpdate}
\begin{algorithmic}[1]
    \renewcommand{\algorithmicrequire}{\textbf{Input:}}
    \renewcommand{\algorithmicensure}{\textbf{Output:}}
    \Require Buffer $\mathcal{B}$, current data $(\mathbf{X}_t, \mathbf{Y}_t)$, current time step $t$

    \If{$|\mathcal{B}| \leq n_B$} 
        \State $\mathcal{B} \gets \mathcal{B} \cup \{(\mathbf{X}_t, \mathbf{Y}_t)\}$ 
    \Else
        \State $i \gets \text{RandInt}([0, t])$
        \If{$i \leq |\mathcal{B}|$}
            \State $\mathcal{B}[i] \gets (\mathbf{X}_t, \mathbf{Y}_t)$
        \EndIf
    \EndIf
    \Ensure  $\mathcal{B}$
\end{algorithmic}
\label{alg: buffer}
\end{algorithm}

\subsubsection*{Local minima exploration}\label{sec: noise}
We examine four sources of noise during deep neural network training in Lines~\ref{line: alg1 explore start}-\ref{line: alg1 explore end} of Algorithm~\ref{alg: driftnet} and Algorithm~\ref{alg: explore}: batch sampling randomness, node dropout~\cite{srivastava2014dropout}, additive gradient noise~\cite{neelakantan2015adding}, and additive input noise~\cite{maharana2022review}. Batch-sampling randomness occurs for all noise types as a subset of data is sampled to compute the gradient, reducing computational load in deep learning experiments. Node dropout, intrinsic to neural networks with dropout layers, involves randomly zeroing out certain parameters within the network. Additive gradient noise introduces Gaussian white noise into the gradient calculation, while additive input noise injects Gaussian white noise into the input data.

\subsubsection*{Information encoding}\label{sec: encode}
When task identities are known during training, an approach to encode the increasing local minima of diverse tasks is to store a limited number of local minima per task. This prevents overwriting learned knowledge. However, when task identities are unknown, DriftNet must efficiently store an increasing number of local minima without overwriting previously learned knowledge, detailed in Lines~\ref{line: alg1 enc start}-\ref{line: alg1 enc end} of Algorithm~\ref{alg: driftnet} and Algorithm~\ref{alg: encoding}.

To group the stored minima, we use the insight that diverse local minima from the same task all perform well on the learned task but tend to perform differently on unlearned tasks. 
Specifically, we maintain a reservoir buffer~\cite{vitter1985random} $\mathcal{B} = {(\mathbf{X}_1,\mathbf{Y}_1), \ldots, (\mathbf{X}_n,\mathbf{Y}_n)}$ of size $n$ to store previously seen data, where $\mathbf{X}_1, \dots, \mathbf{X}_n$ are inputs and $\mathbf{Y}_1, \ldots, \mathbf{Y}_n$ are labels~(Algorithm~\ref{alg: buffer}). It continuously stores data to ensure that each input and label pair has the same probability of being included in the buffer, which is consistent with replay-based lifelong learning approaches~\cite{rolnick2019experience}.
Suppose that the knowledge base $\mathcal{K}$ contains local minima $M_1,\ldots, M_{m(t)}$ at time $t$. We then evaluate the performance vector ($\mathbf{pv}_i, i\in[m(t)]$) of every local minimum in the buffer, defined as $\mathbf{pv}_i \triangleq (pv_{i,1}, \ldots, pv_{i,n})$, where \[
pv_{i,j}\triangleq \text{Criterion}(M_i(\mathbf{X}_j), \mathbf{Y}_j),
\] 
for $i = 1, \ldots, m(t)$ and $j = 1, \ldots, n$, and the Criterion function maps from a pair of prediction and output observation to a real number. In this study, we applied the DBSCAN algorithm~\cite{ester1996density} to cluster the performance vectors $\mathbf{pv}_1, \ldots, \mathbf{pv}_t$, which does not require a pre-specified number of groups.

To further reduce memory cost, one can decrease the memory usage of each local minimum or trainable parameters during training, using parameter-efficient fine tuning (PEFT) strategies such as Adapter~\cite{houlsby2019parameter, pfeiffer2020adapterhub} that trains small adapters and LoRA~\cite{hu2021lora} that limits parameter updates to low-rank matrices. Additionally, the number of local minima within each group is limited to ensure that the knowledge base is not monopolized by a single task, even if it appears frequently.

\subsubsection*{Retrieval via uncertainty quantification}\label{sec: uncertainty}
During the retrieval stage, DriftNet selects the group of local minima from the knowledge base that exhibit minimal output uncertainty (see Lines~\ref{line: alg1 retrieve start}--\ref{line: alg1 retrieve end} of Algorithm~\ref{alg: driftnet} and Algorithm~\ref{alg: retrieve}), as explained below.

The knowledge base contains $\max{(\textbf{gr})}$ groups of local minima, $\mathcal{K}_1, \ldots, \mathcal{K}_{\max{(\textbf{gr})}}$. For a test input $\mathbf{X}_{\text{test}} \in \mathbb{R}^{p}$, the local minima models in the $\mathcal{C}_i$ output predictions $\{\hat{\mathbf{Y}}_{i,1}, \ldots, \hat{\mathbf{Y}}_{i,|\mathcal{K}_i|}\}\triangleq \{M_{\tilde{\theta}}(\mathbf{X}_{\text{test}}): \tilde{\theta}\in \mathcal{K}_i\}$, for $i\in[\max{(\textbf{gr})}]$. The group that minimizes an uncertainty measure is selected to produce the final predictions.

For classification problems, where $\hat{\mathbf{Y}}_{i,j} \in \mathbb{R}^{c}$ and $c$ is the number of classes, we consider three types of uncertainty measures:
\begin{enumerate}
    \item \textbf{Entropy}: Measures the entropy of the averaged output predictions.
    \[
    \operatorname{Entropy}\left(\frac{1}{\tilde{n}_i} \sum_{j=1}^{\tilde{n}_i} \hat{\mathbf{Y}}_{i,j}\right).
    \]
    
    \item \textbf{Variance of hard labels}: Measures the variance of the most likely class labels (hard labels).
    \[
    \operatorname{Var}(\tilde{\mathbf{Y}}_{i,1}, \ldots, \tilde{\mathbf{Y}}_{i,\tilde{n}_i}),
    \]
    where the hard label $\tilde{\mathbf{Y}}_{i,j} \triangleq \argmax(\hat{\mathbf{Y}}_{i,j})$ is the dimension with the largest component value, for $j \in [\tilde{n}_i]$.
    
    \item \textbf{Variance of Soft Labels}: Measures the (trace of) variance of the predicted probability distributions.
    \[
    \text{trace}(\hat{V}_i),
    \]
    where the variance matrix $\hat{V}_i \triangleq \operatorname{Var}(\{\hat{\mathbf{Y}}_{i,1}, \ldots, \hat{\mathbf{Y}}_{i,\tilde{n}_i}\}) \in \mathbb{R}^{c \times c}$, and $\text{trace}$ returns the sum of diagonal elements.
\end{enumerate}

 \subsection*{Measures of drifting features}\label{sec: feature}
\paragraph{Feature}
Consider a network, represented as $A_{\alpha} \circ B_{\beta}$, 
where
$B_{\beta}: \mathbb R^p\rightarrow\mathbb R^q$, 
$A_{\alpha}: \mathbb R^q\rightarrow\mathbb R^c$, and $p, q, c \in \mathbb{N}$. 
For $\mathbf{X} \in \mathbb{R}^{p}$, let $\mathbf{z}  \triangleq B_{\beta}(\mathbf{X})$ denote the output of $B_\beta$.
The vector $\mathbf{z}$ is termed the feature of the network.

\paragraph{Drift}
Suppose that the weights $(\alpha, \beta)$ of the network during training follow a trajectory $(\alpha_1, \beta_1), \ldots, (\alpha_T, \beta_T)$, where $T \in \mathbb{N}$. Assuming that the training loss stabilizes at time $t_0$, the drift of features for an input $\mathbf{X}$ is defined as \[
\text{Drift}_t(\mathbf{X}) \triangleq \mathbf{z}_t - \mathbf{z}_{t_0} = B_{\beta_t}(\mathbf{X}) - B_{\beta_{t_0}}(\mathbf{X}),
\]
for $t = t_0 + 1, \ldots, T$. For image datasets and simulated datasets, $t_0$ is set to $T/2$ and $T/4$, respectively.

\paragraph{Manifold of drifts}
The manifold of drifts represents the space of drift changes, specifically the space spanned by $\text{Drift}_{t_0+1}(\mathbf{X}),\ldots,\text{Drift}_{T}(\mathbf{X})$ for all test inputs $\mathbf{X}$. To account for varying norms, we first normalize the drifts: 
\[
\text{Drift}'_t(\mathbf{X})\triangleq \frac{\text{Drift}_t(\mathbf{X})}{\|\text{Drift}_{t_0}(\mathbf{X})\|_2},
\]

where $\|\cdot\|_2$ denotes the Euclidean $l_2$ norm. 
We then perform the principal component analysis (PCA)~\cite{pearson1901liii} on all normalized drifts: 
\[\{\text{Drift}'_t(\mathbf{X}):t>t_0, \mathbf{X}\in \mathcal{X}_{\text{test}}\},\]
where $\mathcal{X}_{\text{test}}$ is the set of all test inputs. Based on PCA, we obtain $q$ principal components (PCs) $\mathbf{v}_1,\ldots, \mathbf{v}_q$ with corresponding eigenvalues $s_1,\ldots,s_q$ ($s_1\geq \ldots\geq s_q\geq 0$). The effective dimension $d$ of the manifold $\mathcal{M}_{d}$ is defined as
\[d\triangleq \left\lceil \frac{(\sum_{i=1}^q s_i)^2}{\sum_{i=1}^q s_i^2}\right\rceil\in [1,q].\]

\paragraph{Drift rate}
The drift rate quantifies the severity of feature drifts by measuring the variance of the drifts projected onto an estimated manifold. The projection step aims to remove fluctuations orthogonal to the drift manifold. We first project the feature drifts onto the first $d$ dimensions of the PCA components, where $d$ is the effective dimension:
\[
\begin{aligned}
    \text{DriftProj}_t(\mathbf{X}) \triangleq & \mathbb{P}_{\mathcal{M}_{d}} \text{Drift}_t(\mathbf{X}) \\
    = & V_d (V_d^\top V_d)^{-1} V_d^\top \text{Drift}_t(\mathbf{X}),
\end{aligned}
\]
where $V_d \triangleq (v_1, \ldots, v_d) \in \mathbb{R}^{q \times d}$ is the matrix of the first $d$ principal components, and $\mathbb{P}_{\mathcal{M}_{d}} \triangleq V_d (V_d^\top V_d)^{-1} V_d^\top$ is the projection operator onto the subspace spanned by the first $d$ principal components.

The drift rate is then defined as the average of $v(\mathbf{X}_{\text{test}})$ over all test inputs $\mathbf{X}_{\text{test}}$ from the test datasets, where $v(\mathbf{X}_{\text{test}})$ is defined as
\[
v(\mathbf{X}_{\text{test}}) \triangleq \text{trace}(\tilde{V}(\mathbf{X}_{\text{test}})),
\]
where $\tilde{V}(\mathbf{X}_{\text{test}}) \in \mathbb{R}^{d \times d}$ is the sample covariance matrix calculated from $\text{DriftProj}_{t_0+1}(\mathbf{X}), \ldots, \text{DriftProj}_{t_{\text{end}}}(\mathbf{X})$, and $t_{\text{end}}$ is the end timestamp (end of the corresponding training task).

\subsection*{Lifelong learning baselines}\label{sec: baseline}

In this section, we briefly introduce the comparison baselines. The titles provided are the abbreviations used throughout this paper.

\paragraph{Fine-tune}
The Fine-tune baseline continuously updates a single model using gradient descent in the current batch. This method does not retain information from previous tasks and relies solely on the current data. As a result, it provides a naive baseline, highlighting the model's performance without any mechanisms for retaining or recalling past knowledge.

\paragraph{Joint}
The Joint baseline trains a single network on the combined dataset of all tasks, treating it as a single large task. This approach requires access to all data simultaneously, which is often not feasible in real-world scenarios. 
It serves as a performance benchmark, representing the upper limit of what could be achieved when test task identity is unknown with offline data and perfect memory.

\paragraph{Stable}
The Stable baseline continuously trains a network and stores the model at fixed intervals of time. During testing, the final output is obtained by averaging the outputs of the latest $10$ stored models. This approach aims to simulate a scenario where encoding and retrieval are performed without being guided by drift.

\paragraph{Theoretical Limits}
The Theoretical Limits strategy maintains a set of distinct models, one for each task. During both training and inference, the task identity is explicitly provided, allowing the model to use the corresponding task-specific network for label prediction. It is unfeasible in practice and serves as an upper bound for performance, demonstrating the best possible outcomes when task identities are known.

\paragraph{Gen}
Generative classifiers (Gen)~\cite{van2021class} maintain a set of tuples, each consisting of a classifier and a generator, such as a Variational Autoencoder (VAE). The model has one tuple of generator and classifier per task. During training, the task identity is provided and the corresponding tuple is updated based on the current batch. In the inference phase, the generators assist in selecting the appropriate classifier. The classifier associated with the generator that produces the lowest loss in inputs is chosen, using generative models to enhance classification~performance.

\paragraph{ER}
Experience Replay (ER)~\cite{chaudhry2019tiny} is an effective baseline in lifelong learning that operates without the need for known task identities. ER uses Reservoir Sampling~\cite{vitter1985random} to maintain a buffer, ensuring that each data point is stored with equal probability. During training, the model is updated by integrating the current data batch with a batch sampled from this buffer. This approach helps mitigate catastrophic forgetting by revisiting past experiences during the learning~process.

\subsection*{Simulation}\label{sec: toy}

We generated simulated datasets for two linear regression tasks with inputs $(x_1,x_2,x_3)$ and labels $y=\beta_0^*+\beta_1^* x_1+\beta_2^* x_2+\beta_3^* x_3+\varepsilon$, where $\varepsilon \sim \mathcal{N}(0,0.01)$. Specifically, the two tasks were as follows: (1) for the first task, $x_2=0$, $(x_1,x_3) \sim \mathcal{N}(0,I_2)$, and $(\beta_0^*,\beta_1^*,\beta_2^*,\beta_3^*) = (0,1,1,1)$. (2)  for the second task, $x_2=x_1$, $(x_1,x_3) \sim \mathcal{N}(0,I_2)$, and $(\beta_0^*,\beta_1^*,\beta_2^*,\beta_3^*) = (0,-1,-1,-1)$. Therefore, the theoretical manifolds of local minima for the two tasks are ${(0,c,1,1) : c \in \mathbb{R}}$ and ${(0,c,-2-c,-1) : c \in \mathbb{R}}$, respectively. Each task contained $10,000$ samples and was iterated $100$ times ($100$ epochs per task) with batch size $16$. 

We trained DriftNet using the model $y=\beta_0+\beta_1 x_1+\beta_2 x_2+\beta_3 x_3$, with stochastic gradient descent (SGD) with a learning rate of $0.001$. Gaussian noise with mean $0$ and variance $\sigma^2I_4$ was added to each gradient during training, similarly to theoretical proposals for representational drift~\cite{aitken2022geometry,qin2023coordinated,pashakhanloo2023stochastic} and stochastic Langevin gradient descent~\cite{teh2016consistency}.
We maintained a buffer storing 10 randomly picked batches for every learned task, and then evaluated performance vectors of all stored weights every 10 epochs{, with an epoch being a single pass through the entire training dataset.}
The network parameters were stored at the end of each epoch. A buffer was maintained, which stored $10$ random batches per task. Every $10$ epochs, each stored minimum was evaluated in the buffer using a $0$-$1$ metric, where a value of $1$ indicated that the squared test loss exceeded three times the standard deviation of $\varepsilon$ ($0.03$). The DBSCAN algorithm, with cosine distance and a hyperparameter $\epsilon=1$, was applied to group the stored local minima.  During retrieval, given a test input, the variance of the outputs within the same group was regarded uncertainty. The group with minimal uncertainty was selected to provide the output. In contrast, for the stable baseline as a control, the weights were stored similarly at each epoch, and the average output from all stored weights was used for any test inputs. This baseline approach did not involve noise injection during training and served to highlight the differences in performance and robustness compared to DriftNet. 
Experiments were conducted with 50 repetitions for the simulated datasets.

\subsection*{Image classification}%
For approaches training a single model, including Fine-tune, Joint, and ER, we employed ResNet-18~\cite{he2016deep} with $64$ initial filters. For approaches involving multiple models, including Theoretical Limits, Stable, Gen, and DriftNet, we used a reduced version of ResNet-18 with initial filters $20$, as in work~\cite{lopez2017gradient, chaudhry2019continual}. Furthermore, we used a CNN-based VAE with two $3\times 3$ convolutional layers in the encoders, following the design in work~\cite{lee2020neural}. For all methods and datasets, the training batch size was set to $16$. We used the AdamW optimizer~\cite{loshchilov2017decoupled} with a learning rate of $0.001$, betas of $(0.9, 0.999)$, and a weight decay of $0.01$.

For DriftNet, we explored various sources of noise: dropout noise with a probability of $0.5$,  input noise by injecting Gaussian noise with a standard deviation of $0.01$ into the input images, and gradient noise by injecting Gaussian noise to the gradient of every layer $g \in \mathbb{R}^{l}$ with a standard deviation:
${0.1 \| g \|_2}/{\sqrt{l}}$.
The network parameters were saved every five epochs. A reservoir buffer~\cite{vitter1985random} of size $500$ was maintained over time to store previous data samples uniformly. We applied the DBSCAN algorithm to group all stored local minima, using a hyperparameter $\epsilon=0.5$ (the maximum distance between two samples to be considered neighbors) and the cosine distance for the grouping. During testing, the batch size was maintained at 16, identical to the training batch size. ER used a reservoir buffer of size $2000$ for CIFAR-10 and $5000$ for CIFAR-100, respectively.

\subsection*{Natural language processing}\label{sec: nlp}
We selected four datasets for our experiments, sequentially learning from each dataset: AG's News, Amazon Review Full, DBpedia, and Yahoo! Answers~\cite{zhang2015character}:
\begin{itemize}
\item \textbf{AG's News:} This dataset contains $496,835$ categorized news articles classified into four largest classes: World, Sports, Business, and Science/Technology. The number of training samples for each class is $30,000$, with $1,900$ testing samples per class.
\item \textbf{Amazon Review Full:} This sentiment analysis dataset contains reviews on products with ratings from $1$ to $5$ stars. It includes $600,000$ training samples and $130,000$ testing samples per class for the full score prediction.
\item \textbf{DBpedia:} This dataset is a community effort to extract structured information from Wikipedia. It includes $14$ nonoverlapping classes from DBpedia 2014, such as Animal, Plant, Album, and Film. Each class has 40,000 training samples and 5,000 testing samples.
\item \textbf{Yahoo! Answers:} This topic classification dataset contains $10$ topics, including Society \& Culture, Science \& Mathematics, and Health. Each class contains $140,000$ training samples and $5,000$ testing samples.
\end{itemize}

In our experiments, we used DriftNet, which trains a parameter-efficient component using Low-Rank Adaptation (LoRA) over a pre-trained LLM GPT-2~\cite{radford2019language}, which consists of $12$ GPT2 blocks.  The hyperparameters for LoRA include the rank $r = 128$, the scaling factor $\alpha = 64$, and the dropout probability $p = 0.5$.

Each task was trained using SGD over $5$ epochs with a batch size of $8$. We employed a cyclic learning rate scheduler, which linearly increased the learning rate from $10^{-5}$ to $10^{-2}$, then decreased back to $10^{-5}$, and repeated this cycle four times per epoch. The network parameters were saved at the lowest learning rate to ensure stability.
To maintain a history of previous data samples, a reservoir buffer~\cite{vitter1985random} of size $500$ was used. This buffer stored samples uniformly over time, allowing for efficient sampling of past data. For grouping the stored local minima, we applied the DBSCAN algorithm with a hyperparameter $\epsilon = 0.5$ and used cosine distance as the metric for grouping.

\clearpage
\section*{Data availability}
{Data used for image classification experiments are available in ref.~\cite{krizhevsky2009learning}.
Data used for natural language processing experiments are available in ref.~\cite{zhang2015character}.
}

\section*{Acknowledgments}
J.L., J.Ding, and N.L. acknowledge the support from the NIH 5R01LM014465.
J.Ding acknowledges the support from the ARO W911NF-23-10315. J.L. acknowledges the support from the AFOSR YIP FA9550-22-1-0228.

\section*{Author contributions}
J.Du, J.L and J.Ding conceived the idea and designed the research. J.Du developed the algorithm, conducted experiments, prepared figures, and wrote the manuscript. X.Z. revised the figures. J.Z., X.Z., H.S., Y.Y. and N.L. provided discussions during the development. X.X. and G.W. contributed to experimental design of language models. All authors contributed to the revising of the manuscript. J.L. and J.Ding supervised the study.

\section*{Competing interests}
All authors have no competing interests.

\section*{Additional information}
\textbf{Correspondence}
and requests for materials should be addressed to Jia Liu or Jie Ding.

\raggedbottom 
\bibliography{sn-bibliography}%

\newpage
\section*{Figures and Figure   Captions}

\begin{figure}[H]
    \centering
    \includegraphics[width=0.98\textwidth]{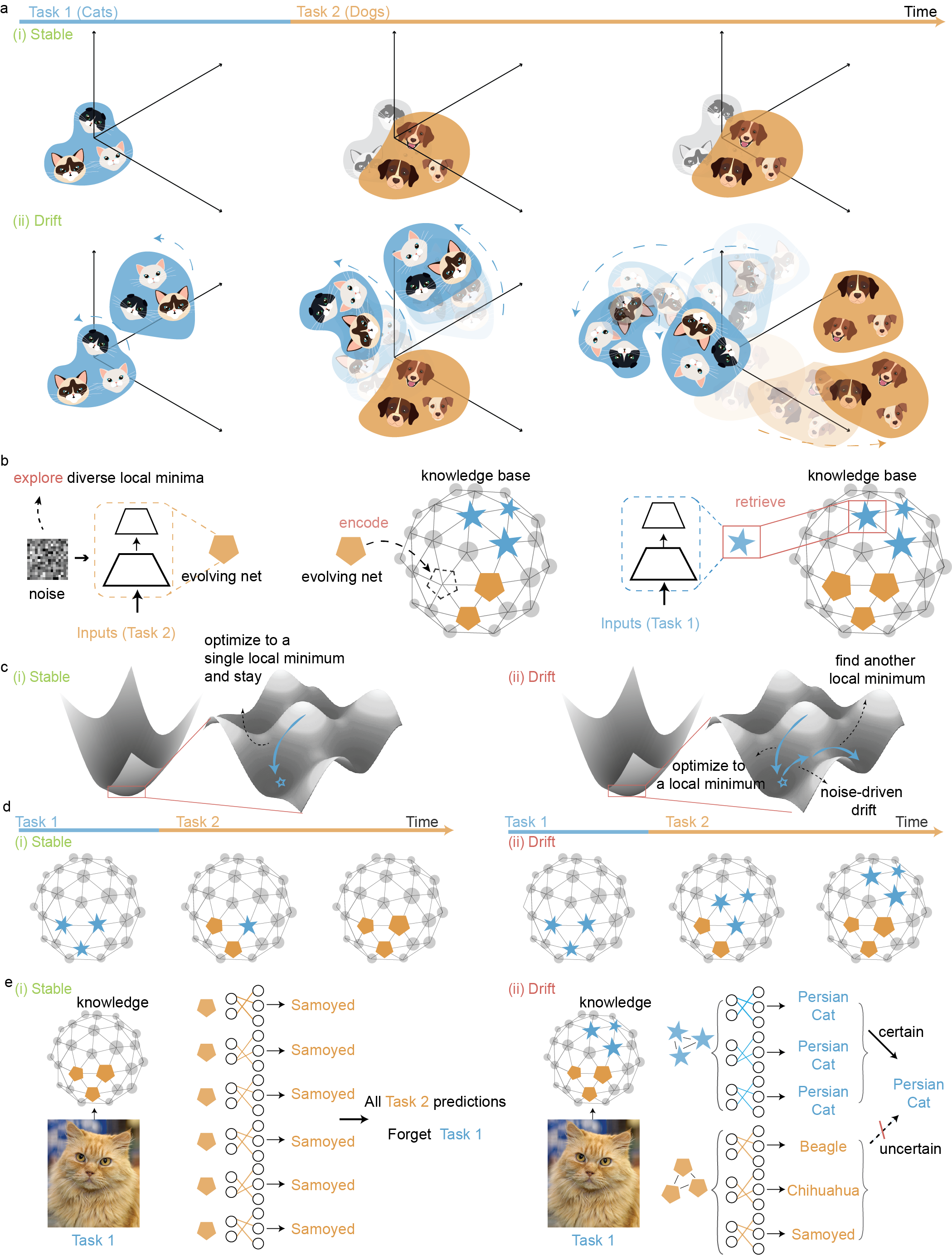}
    \caption*{}
\end{figure}

\clearpage
\begin{figure*}[ht]
    %\internallinenumbers 
    \caption{
    \textbf{Drift-inspired mechanism to prevent catastrophic forgetting in lifelong learning.} 
    \textbf{a}, Schematic illustration of stable and drifting networks. A stable network continuously learns new tasks, but overwrites previously acquired knowledge. In contrast, a drifting network allows its representations to continuously drift, thereby avoiding overwriting when learning new tasks.
   \textbf{b-e}, Implementation of the neural representational drift-inspired lifelong learning algorithm,
    \textbf{b}, Schematic overview showing the three steps of DriftNet: exploration~(left), encoding~(middle), and retrieval~(right). DriftNet features an evolving model for exploration and a knowledge base for encoding and retrieving grouped task-specific information.
    \textbf{c}, Exploration step. Enabled by external noise, instead of (i) remaining fixed at a single local minimum like a stable network, the drifting network (ii) explores alternate minima in the current loss landscape.
    \textbf{d}, Encoding step. In a stable network, newly learned local minima overwrite previous ones, leading to gradual forgetting of previous tasks. In contrast, DriftNet constantly organizes and groups Task 1-specific local minima in the knowledge base during the learning of Task 2, preventing them from being overwritten by the newly learned minima from Task 2.
    \textbf{e},  Retrieval step. A stable network (i) cannot identify Task 1 information due to the forgetting caused by overwriting. In contrast, DriftNet (ii) can still identify the grouped Task 1-specific information after learning a new task, enabled by its drifting characteristic.
}
    \label{fig: fig1}
\end{figure*}

\clearpage
\begin{figure*}[ht]
    \centering   \includegraphics[width=\textwidth]{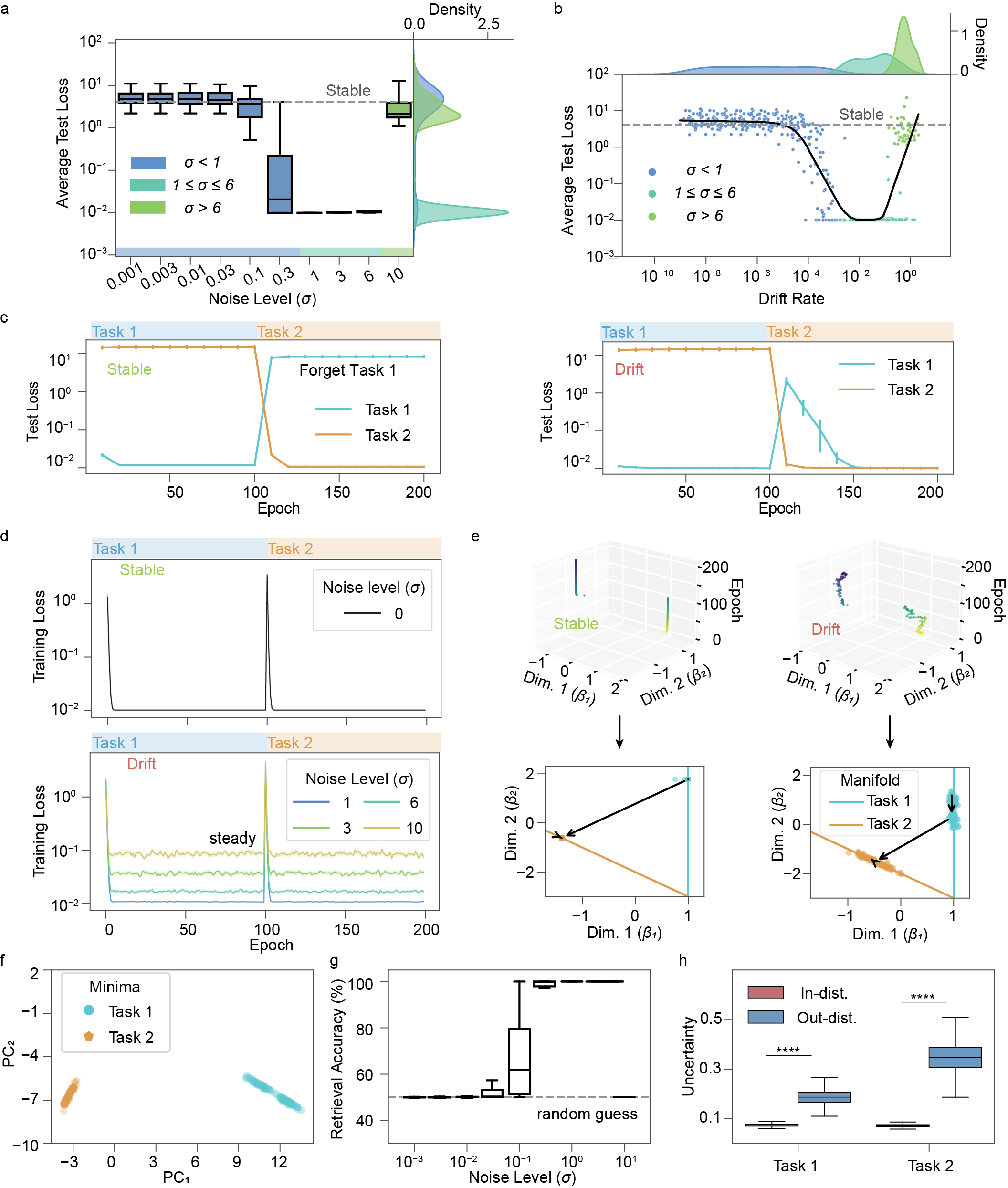}
    \caption*{}
\end{figure*}
\clearpage
\begin{figure*}[ht]
%\internallinenumbers 
        \caption{\textbf{Benchmarking DriftNet lifelong learning performance on simulated datasets.}
    \textbf{a-d}, Statistical results of DriftNet lifelong learning performance on simulated linear regression datasets. $n = 50$ experimental replicates.
    \textbf{a},
    Boxplots with density plots of average test loss, showing the average test loss of two tasks relative to noise scales. The gray dotted line represents the stable baseline.
    \textbf{b},
    Boxplots with density plots of drift rate, showing the average test loss of two tasks relative to noise scales. The gray dotted line represents the stable baseline. The black line represents the locally weighted scatter plot smoothing (LOWESS) curve of the average test loss with a fraction of $0.3$.
    \textbf{c}, 
    Statistical summary of test losses for two tasks relative to epoch for drift (left) and stable (right) networks, respectively. The value represents the mean $\pm$ SE, $n = 50$ experimental replicates.
    \textbf{d},
     Statistical summary of training losses with different noise levels ($\sigma$) relative to the epoch of drift (top) and stable networks (bottom), respectively. The value represents the mean $\pm$ SE, $n = 50$ experimental replicates.
    \textbf{e},  
    Scatter dots plot showing the trajectory of two model weights $(\beta_1,\beta_2)$ over time. The top plot contains points over the epoch, $(\beta_1,\beta_2, \text{epoch})$, in 3d space; and the bottom plot contains $(\beta_1,\beta_2)$ data for all epochs. The orange and blue lines indicate the theoretical minima manifolds of Tasks 1 and 2, respectively.  $\sigma=3$.
    \textbf{f}, 
    Scatter dot plots showing performance vectors of minima mapped onto the first two principal components (PCs).
    \textbf{h},  
    Boxplot showing the retrieval accuracy relative to noise scales ($\sigma$). The dashed green line indicates the stable baseline.
     \textbf{g}, 
    Boxplot showing the uncertainty of task-specific groups of local minima, evaluated on batch of input data of the relevant task (in-distribution), and irrelevant task (out-distribution). $\sigma=0.001$, batch size $16$. Box, $75\%$ and $25\%$ quantiles. Line, median. Whisker, median $\pm$ $1.5\times$ interquartile range (IQR). $n=50$ experimental replicates. ****\textit{p}-values of the Mann-Whitney $U$ test and the student \textit{t}-test are less than $10^{-4}$.
    }
    \label{fig: fig2}
\end{figure*}
\clearpage
\begin{figure*}[ht]
    \centering
    \includegraphics[width=\textwidth]{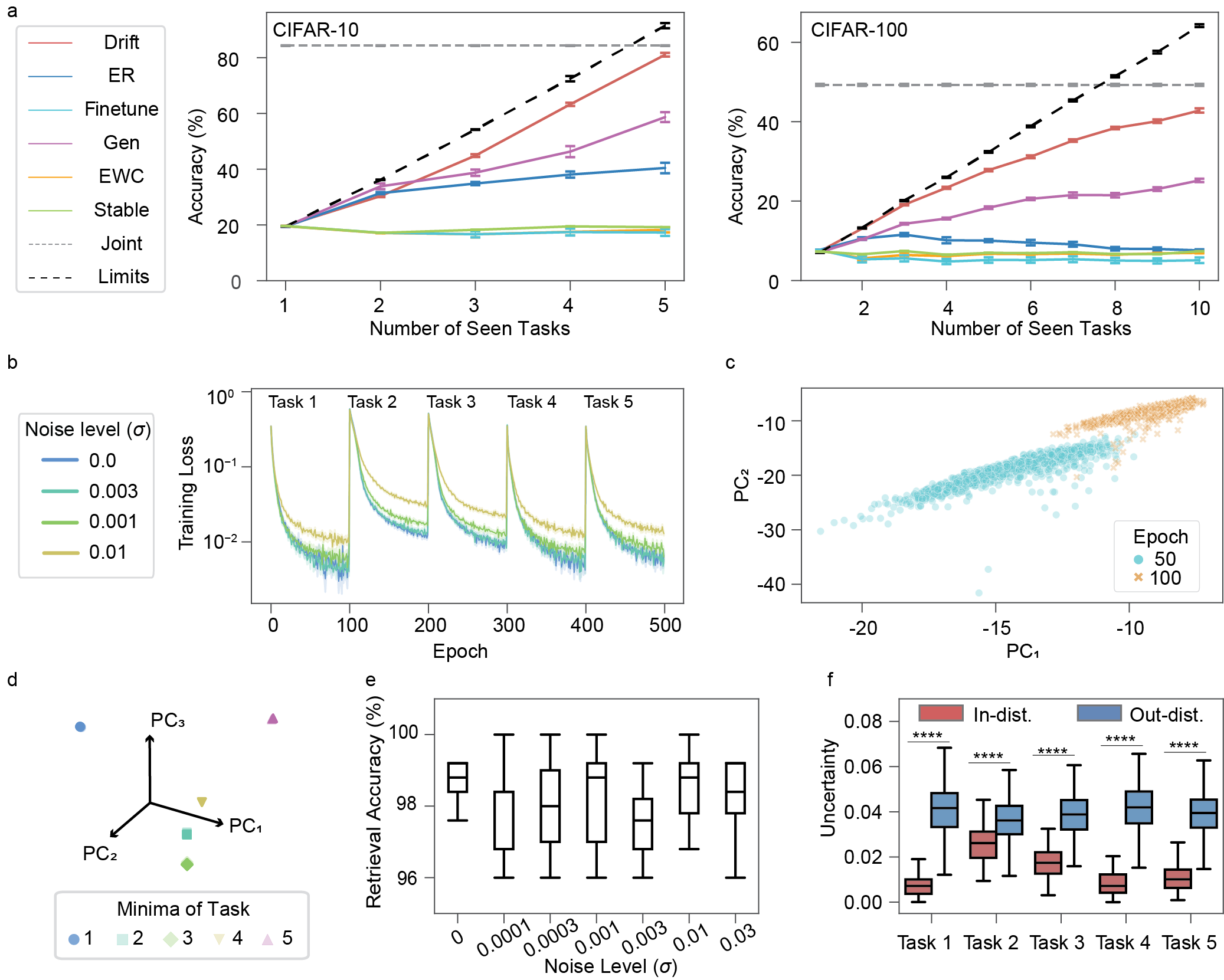}
        %\internallinenumbers
    \captionof{figure}{
    \textbf{Image classification.}
    \textbf{a}, 
    Statistical summary of the average test accuracy of all tasks relative to the number of seen tasks, for CIFAR-10 (left) and CIFAR-100 (right). The value represents the mean $\pm$ SE, $n = 10$ experimental replicates.
    \textbf{b-f}, Statistical results of DriftNet performance on CIFAR-10.
    \textbf{b}, 
    Statistical summary of the training loss with different noise scales ($\sigma$) relative to epoch. The value represents the mean $\pm$ SE, $n = 10$ experimental replicates.
    \textbf{c}, 
    Scatter dots plot showing feature drifts of the first category, projected on the first two principal components (PC) of the feature drifts (see Methods), at epochs $50$ and $100$. $\sigma=0.001$. 
    \textbf{d}, 
     Scatter dots plot showing the performance vectors of the minima, evaluated in the buffer (see Methods), mapped to the first three principal components. $\sigma=0.001$.
     \textbf{e},
    Boxplot showing the retrieval accuracy relative to noise scales ($\sigma$) (see Methods).
    \textbf{f},  
    Boxplot showing the uncertainty of task-specific groups of local minima, evaluated on batch of input data from the relevant task (in-distribution), and irrelevant task (out-distribution), see Methods. $\sigma=0.001$, batch size $16$. %
    Box, $75\%$ and $25\%$ quantiles. Line, median. Whisker, the most extreme data point within the median $\pm$ $1.5\times$ interquartile range (IQR). $n=10$ experimental replicates. ****\textit{p}-values of the Mann-Whitney $U$ test and the student \textit{t}-test are less than $10^{-4}$.
    }
    \label{fig: fig3}
\end{figure*}
\clearpage
\begin{figure*}[ht]
    \centering
    \includegraphics[width=\textwidth]{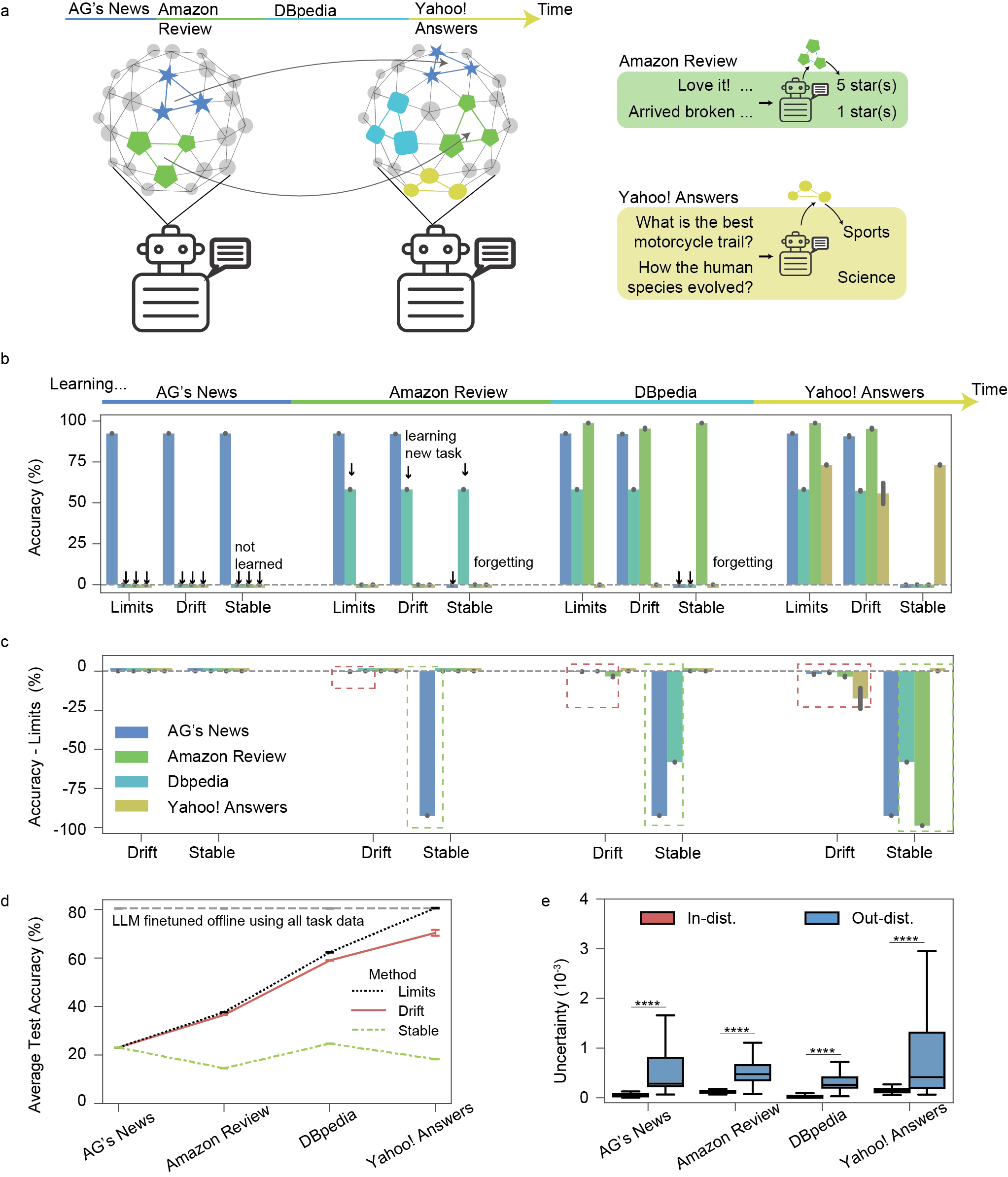}
\end{figure*}
\clearpage
\begin{figure*}[ht]
%\internallinenumbers
    \caption{
    \textbf{Large language model GPT-2 applications.}
    \textbf{a},
    Illustration of the lifelong learning chatbot (DriftNet). After learning four language tasks, the chatbot retrieves the relevant set of minima to provide accurate responses.
    \textbf{b}, 
    Barplots showing the test accuracy for each of four tasks during their learning process. The bar represents the mean $\pm$ SE, $n=5$ experimental replicates.
    \textbf{c},
     Barplots showing the difference in test accuracy between methods and Theoretical Limits for each of the four tasks during the learning process. The bar represents the mean $\pm$ SE, $n=5$ experimental replicates. Red boxes indicate Drift networks learning a new task with mild forgetting. Green boxes indicate Stable networks forgetting previous tasks.
     \textbf{d},
    Statistical summary of the average test accuracy of all tasks relative to the number of seen tasks, for different methods. The value represents the mean $\pm$ SE, $n = 5$ experimental replicates. {The gray dashed line represents the Joint baseline, where a pre-trained large language model is fine-tuned offline using all task data.}
    \textbf{e},
    Boxplots showing the uncertainty of task-specific groups of local minima, evaluated on batch of input data from the relevant task (in-distribution), and irrelevant task (out-of-distribution), see Methods. Batch size $16$. 
    Box, $75\%$ and $25\%$ quantiles. Line, median. Whisker, the most extreme data point within the median $\pm$ $1.5\times$ interquartile range (IQR). $n=5$ experimental replicates.
    }
    \label{fig: fig4}
\end{figure*}

\clearpage

\FloatBarrier
\section*{Extended Data Figures and Extended Data Figure Captions}
\FloatBarrier
\setcounter{figure}{0}
\begin{figure*}[ht]
    \centering
    \includegraphics[width=0.9\textwidth]{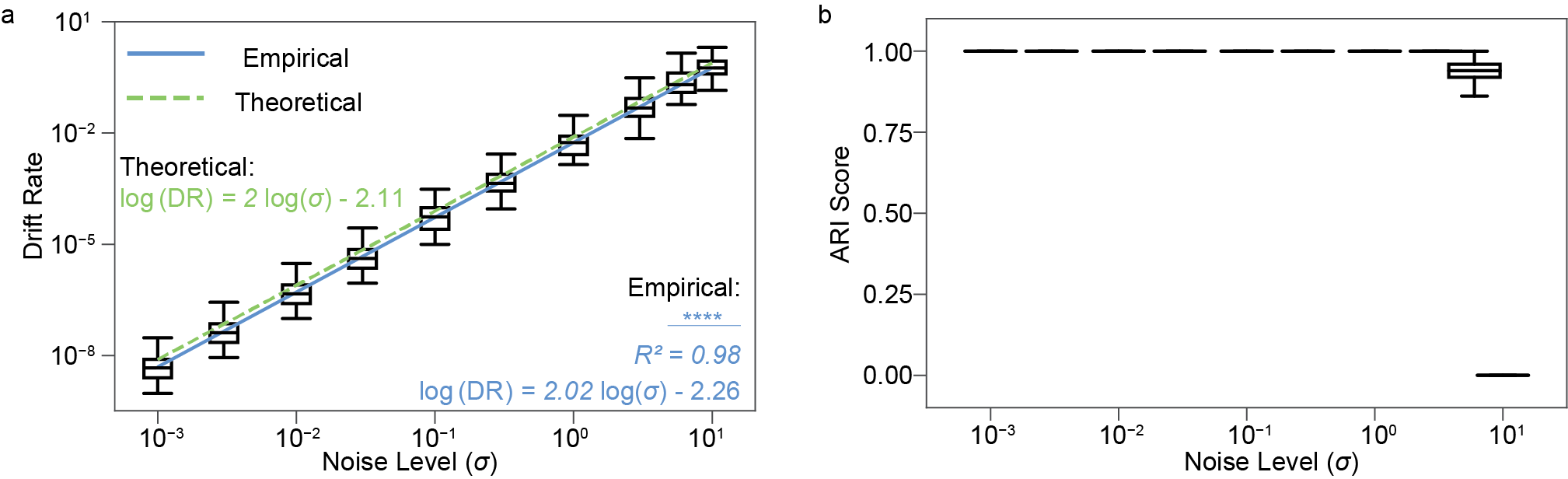}
    \extendcaption{
    %\internallinenumbers
    \textbf{a,} 
    Boxplots showing the drift rates (see Methods) relative to noise levels ($\sigma$). The blue line and green dashed line indicate the empirically fitted and theoretical lines, respectively. ****\textit{p}-value of F test is less than $10^{-4}$.%
    \textbf{b,} 
    Boxplots showing the Adjusted Rand Index (ARI) score (see Methods) that evaluates the grouping precision compared to the ground truth, relative to the noise scales ($\sigma$). 
    Box, $75\%$ and $25\%$ quantiles. Line, median. Whisker, the most extreme data point within the median $\pm$ $1.5\times$ interquartile range (IQR). $n=50$ experimental replicates. 
    }
    \label{fig: fig2_extended}
\end{figure*}

\clearpage
\begin{figure}[H]
    \centering
    \includegraphics[width=\textwidth]{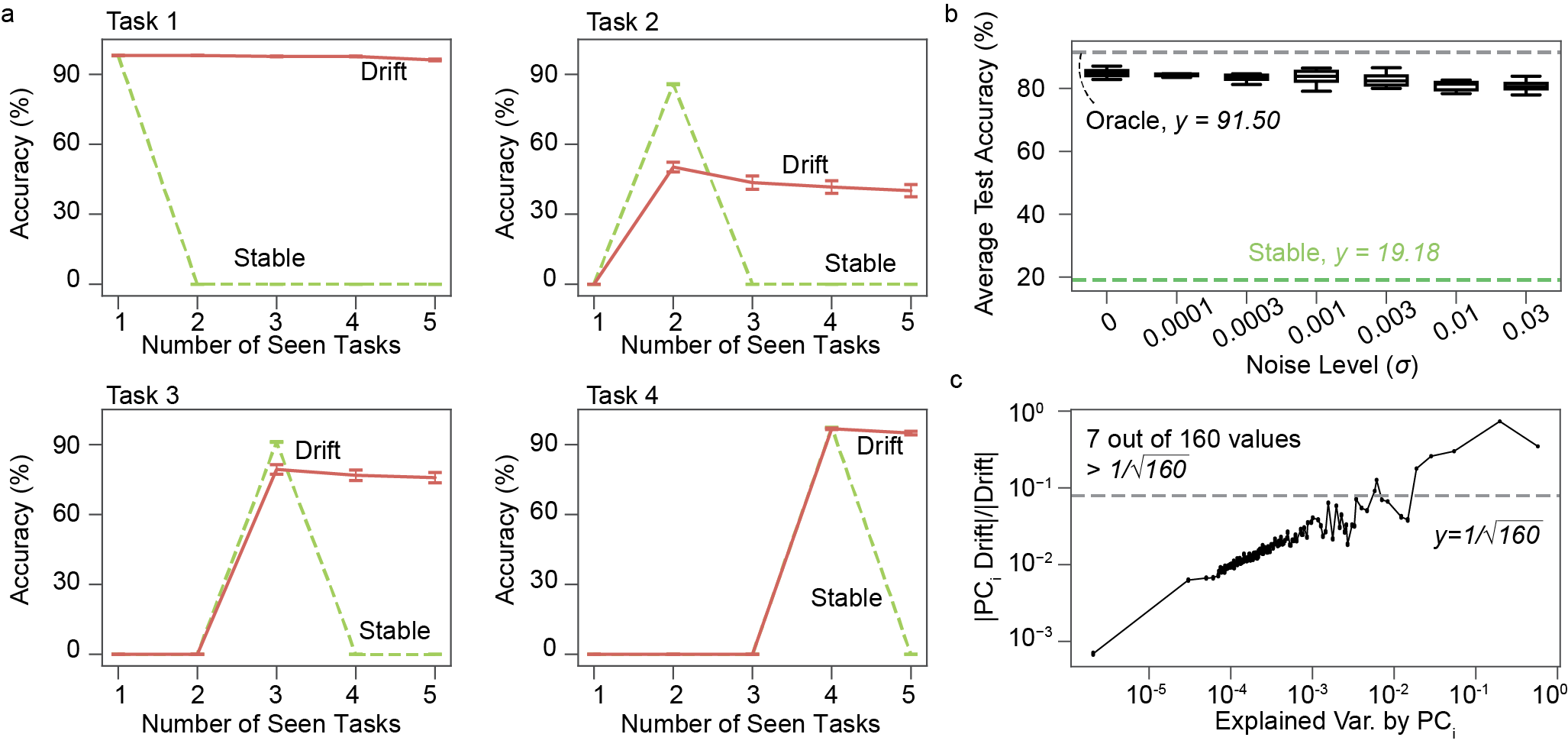}

    \extendcaption{
    %\internallinenumbers
    Statistical comparisons of test performance on CIFAR-10 dataset across different networks.
    \textbf{a,} 
    Statistical summary of the test accuracy of the first four tasks relative to the number of tasks seen for stable and drift networks. The value represents the mean $\pm$ SE, $n = 10$ experimental replicates, $\sigma=0.001$.
    \textbf{b,}
    Boxplots showing the average test performance of all tasks relative to noise scales ($\sigma$).
    \textbf{c,} 
    Ratio of drift in different PC dimensions (PC$_i$, $i=1,\dots, 160$) versus the variance explained by the corresponding PC dimensions (see Methods) of the first category. Value represents mean $\pm$ SE, $n = 1000$ images. The dashed line highlights the reference line. $\sigma=0.001$. %
    }
    \label{fig: fig3_extended}
\end{figure}

\begin{figure*}[ht]
    \centering
    \includegraphics[width=\textwidth]{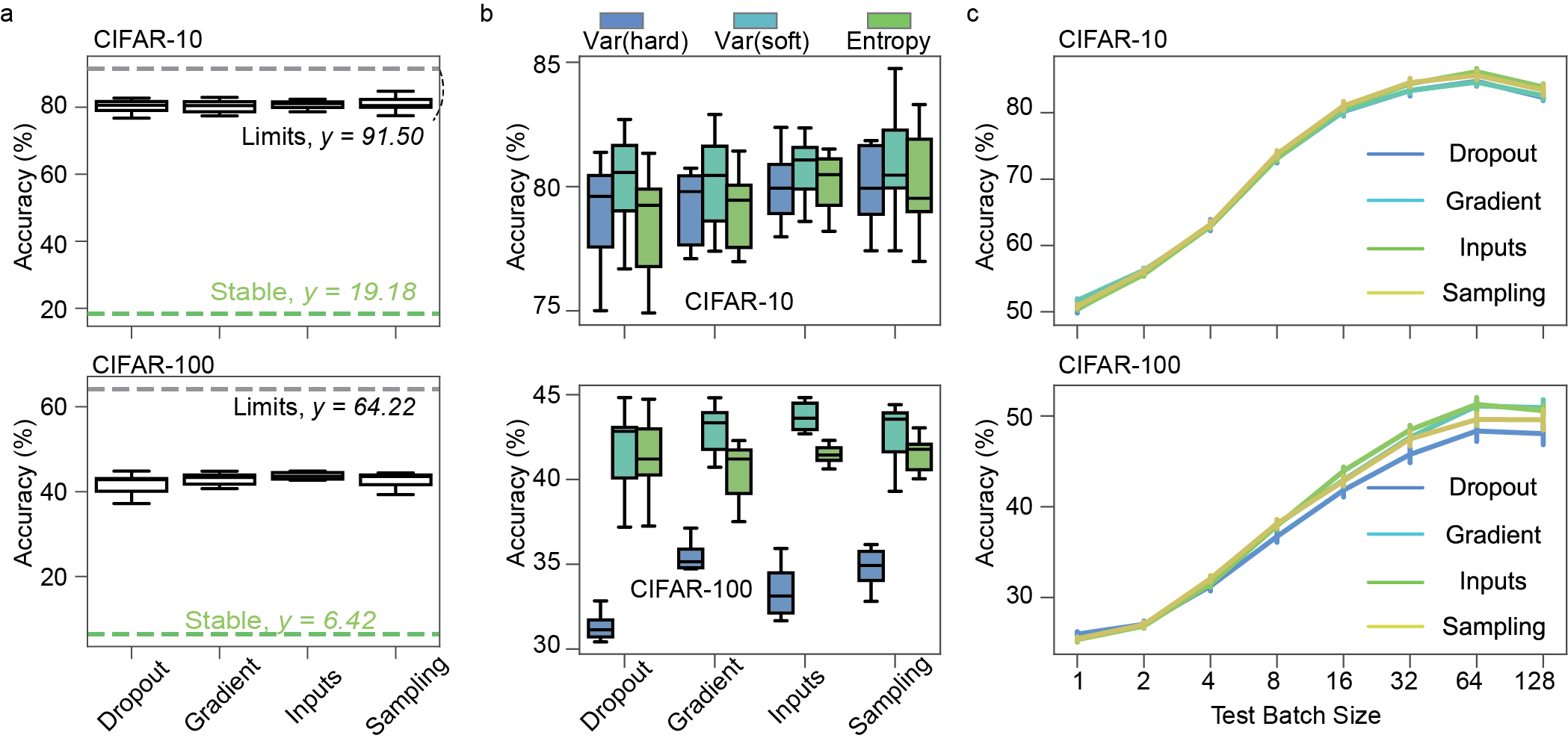}
    \extendcaption{%\internallinenumbers
    Statistical results of DriftNet performance on CIFAR-10 and CIFAR-100 datasets. $n = 10$ experimental replicates. 
    \textbf{a},
   Boxplots showing the average test accuracy of all tasks relative to different noise types, for CIFAR-10 (top) and CIFAR-100 (bottom),  respectively. Gray and green dashed lines indicate the performance from Theoretical Limits and stable network as control. {Pairwise Mann–Whitney \textit{U}-test \textit{p}-values are all above $0.05$.}
    \textbf{b},
    Boxplots showing average test accuracy of all tasks relative to different uncertainty measures (see Methods), for CIFAR-10 (top) and CIFAR-100 (bottom), respectively.  %
    \textbf{c},
    Line plots showing average test accuracy of all tasks relative to different test batch sizes for CIFAR-10 (top) and CIFAR-100 (bottom). The value represents the mean $\pm$ SE, $n = 10$ experimental replicates.
    Box, $75\%$ and $25\%$ quantiles. Line, median. Whisker, the most extreme data point within the median $\pm$ $1.5\times$ interquartile range (IQR). $n=10$ experimental replicates. 
    }
    \label{fig: fig3_extended_2}
\end{figure*}

\begin{figure*}[ht]
    \centering
    \includegraphics[width=\textwidth]{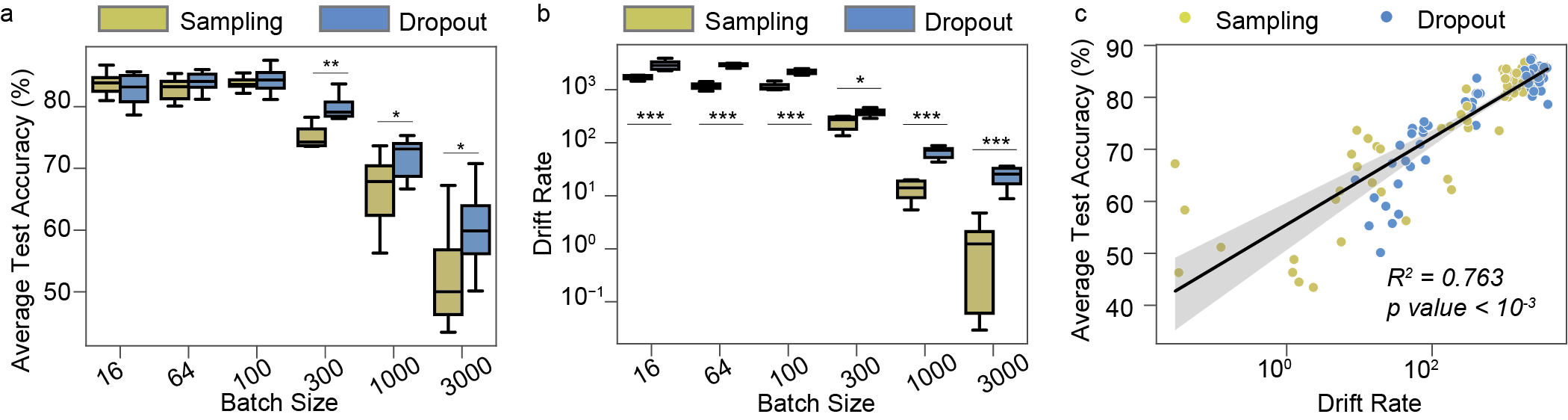}
    \extendcaption{
    %\internallinenumbers
    Statistical summary of the effect of noises on CIFAR-10 dataset with different batch sizes, for networks with batch sampling noise and with dropout, respectively. 
    \textbf{a},
    Boxplots showing the average test accuracy of all tasks relative to different training batch sizes, for networks with batch sampling noise and with dropout, respectively. {Mann–Whitney \textit{U}-test results: $P=0.571$ at $16$, $P=0.186$ at  $64$, $P=0.345$ at $100$, **$P=0.006$ at $300$, *$P=0.021$ at $1000$, *$P=0.021$ at $3000$.}
    \textbf{b},
    Boxplots showing the drift rate relative to different training batch sizes, for networks with batch sampling noise and with dropout, respectively. {Mann–Whitney \textit{U}-test results: ***$P=3.30\times 10^{-4}$ at $16$, ***$P=1.83\times 10^{-4}$ at $64$, ***$P=1.83\times 10^{-4}$ at $100$, ***$P=1.73\times 10^{-2}$ at $300$, *$P=2.46\times 10^{-4}$ at $1000$, ***$P=1.83\times 10^{-4}$ at $3000$.}
    \textbf{c},
    Scatter dots and line plot showing the average test accuracy of all tasks relative to different drift rates, for networks with batch sampling noise and with dropout, respectively. The dots represent individual experimental values, and the line represents the fitted linear regression line of the drift rate relative to the logarithm of drift rate with 95\% confidence interval (shaded area). Box, $75\%$ and $25\%$ quantiles. Line, median. Whisker, the most extreme data point within the median $\pm$ $1.5\times$ interquartile range (IQR). $n=10$ experimental replicates. 
    }
    \label{fig: fig3_extended_3}
\end{figure*}

\begin{figure*}[ht]
    \centering
    \includegraphics[width=\textwidth]{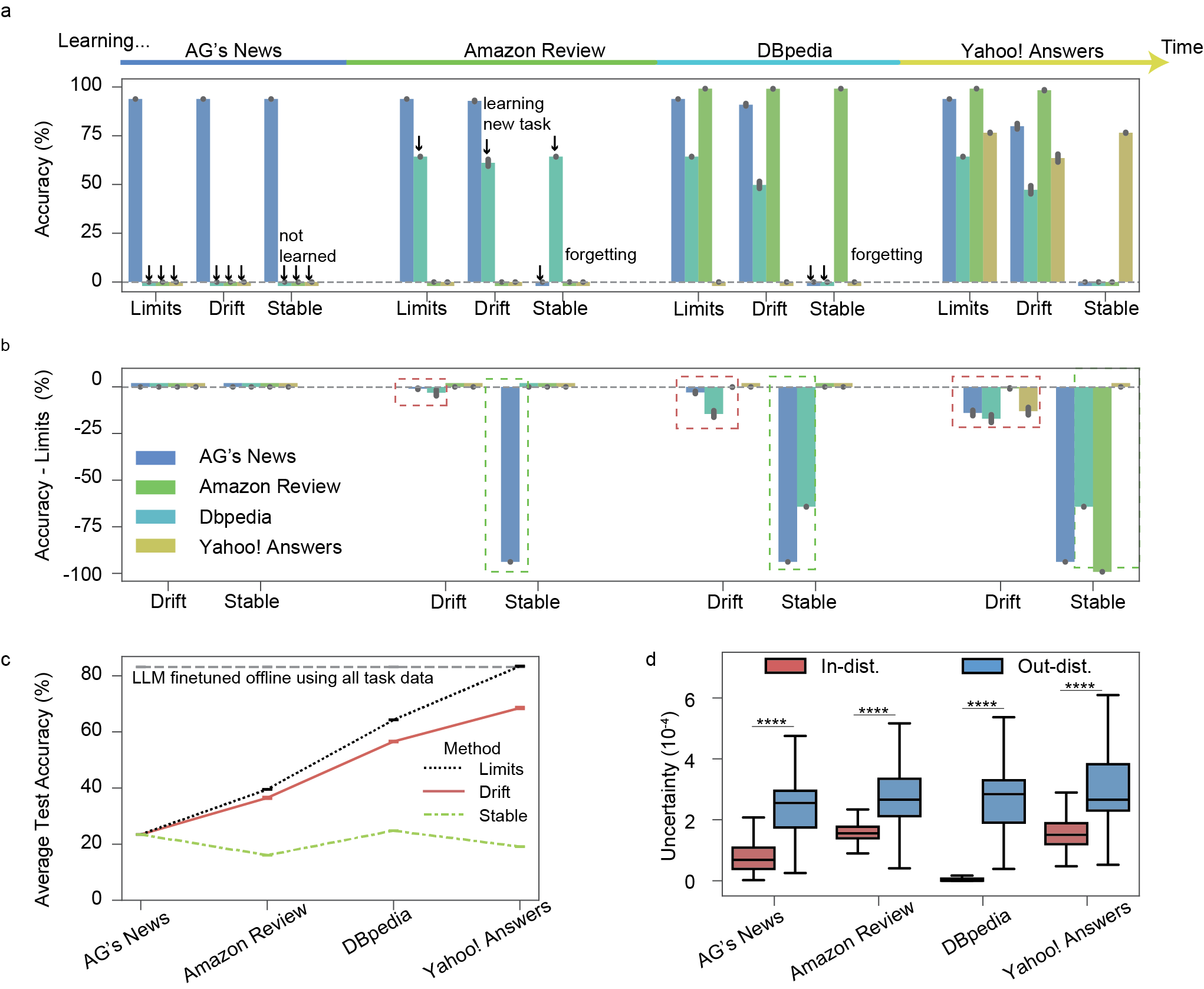}
    \extendcaption{
    %\internallinenumbers
    \textbf{Large language model RoBERTa-Large applications  }
   \textbf{a}, 
    Barplots showing the test accuracy for each of four tasks during their learning process. The bar represents the mean $\pm$ SE, $n=5$ experimental replicates.
    \textbf{b},
     Barplots showing the difference in test accuracy between methods and Theoretical Limits for each of the four tasks during the learning process. Bar represents mean $\pm$ SE, $n=5$ experimental replicates. {Red boxes indicate Drift networks learning a new task with mild forgetting. Green boxes indicate Stable networks forgetting previous tasks.}
     \textbf{c},
    Statistical summary of the average test accuracy of all tasks relative to the number of seen tasks, for different methods. The value represents the mean $\pm$ SE, $n = 5$ experimental replicates. {Gray dashed line represents the Joint baseline, where a pre-trained large language model is fine-tuned offline using all task data.}
    \textbf{d},
    Boxplots showing the uncertainty of task-specific groups of local minima, evaluated on batch of input data from the relevant task (in-distribution), and irrelevant task (out-of-distribution), see Methods. Batch size $16$. %
    Box, $75\%$ and $25\%$ quantiles. Line, median. Whisker, the most extreme data point within the median $\pm$ $1.5\times$ interquartile range (IQR). $n=5$ experimental replicates. {****\textit{p}-values of the Mann-Whitney $U$ test and student \textit{t}-test are less than $10^{-4}$.}
    }
    \label{fig: fig4_extended}
\end{figure*}

\end{document}